\documentclass[lettersize,journal,compsoc]{IEEEtran}
\usepackage{amsmath,amsfonts,amssymb}
\usepackage{algorithm}
\usepackage{algpseudocode}
\usepackage{array}
\usepackage[caption=false,font=normalsize,labelfont=sf,textfont=sf]{subfig}
\usepackage[bookmarks=true, hidelinks, breaklinks,colorlinks,citecolor=blue]{hyperref}
\usepackage{textcomp}
\usepackage{stfloats}
\usepackage{url}
\usepackage{verbatim}
\usepackage{graphicx}
\usepackage{cite}
\usepackage[table]{xcolor}
\usepackage{booktabs}
\usepackage{multirow}
\usepackage{makecell}
\usepackage{enumitem}
\usepackage{tcolorbox}
\usepackage{colortbl}
\usepackage{tabularx}
\usepackage{ragged2e}
\newcolumntype{Y}{>{\raggedright\arraybackslash}X}
\newcolumntype{L}[1]{>{\raggedright\arraybackslash}p{#1}}
\newcolumntype{P}[1]{>{\centering\arraybackslash}p{#1}}
\newcolumntype{C}{>{\centering\arraybackslash}X}
\makeatletter
\floatevery{algorithm}{\def\@fs@post{\kern0pt\hrule\kern-6mm\relax}}
\long\def\@makecaption#1#2{%
\ifx\@captype\@IEEEtablestring%
\footnotesize\bgroup\par
\@IEEEtabletopskipstrut{\normalfont\sffamily\footnotesize\justifying\noindent #1.\nobreakspace #2\par}%
\addvspace{0.5\baselineskip}\egroup%
\@IEEEtablecaptionsepspace
\else
\@IEEEfigurecaptionsepspace
\setbox\@tempboxa\hbox{\normalfont\sffamily\footnotesize {#1.}\nobreakspace #2}%
\ifdim \wd\@tempboxa >\hsize%
\setbox\@tempboxa\hbox{\normalfont\sffamily\footnotesize {#1.}\nobreakspace}%
\parbox[t]{\hsize}{\normalfont\sffamily\footnotesize \noindent\unhbox\@tempboxa#2}%
\else%
\hbox to\hsize{\normalfont\sffamily\footnotesize\box\@tempboxa\hfil}%
\fi\fi}
\makeatother

\let\citep\cite
\let\citet\cite

\begin{document}

\title{Hybrid Token Compression for Vision-Language Models
}

\author{Jusheng Zhang$^\dagger$, Xiaoyang Guo$^\dagger$, Tongyu Mo$^\dagger$, Qinhan Lv,
Wenhao Chai, Jian Wang, \\ Wenhao Wang$^*$, Keze Wang$^*$, IEEE Member, and Liang Lin$^*$, IEEE Fellow 
\IEEEcompsocitemizethanks{
\IEEEcompsocthanksitem Jusheng Zhang, Xiaoyang Guo, Tongyu Mo, Qinhan Lv, Keze Wang, and Liang Lin are with Sun Yat-sen University.
\IEEEcompsocthanksitem Wenhao Chai is with Princeton University.
\IEEEcompsocthanksitem Jian Wang is with Snap Inc.
\IEEEcompsocthanksitem Wenhao Wang is with Vast Intelligence Lab.\\}
\thanks{$^\dagger$ Jusheng Zhang, Xiaoyang Guo, and Tongyu Mo contributed equally.}
\thanks{$^*$ Wenhao Wang, Keze Wang, and Liang Lin are the corresponding authors.}
}

\markboth{}{}

\IEEEtitleabstractindextext{%
\begin{abstract}
\justifying
    Vision-language models (VLMs) have transformed multimodal reasoning, but feeding hundreds of visual patch tokens to LLMs incurs quadratic computational costs, straining memory and context windows. Traditional approaches face a trade-off: continuous compression dilutes high-level semantics like object identities, while discrete quantization loses granular details such as textures. We challenge this by introducing \textbf{HTC-VLM}, a hybrid framework that disentangles semantics and appearance through dual channels, \textit{i.e.}, a continuous pathway for fine-grained details via ViT patches and a discrete pathway for symbolic anchors using MGVQ quantization projected to four tokens. These are fused into a 580-token hybrid sequence and compressed to one token via a disentanglement attention mask and \texttt{<voco>} bottleneck, ensuring efficient, grounded representations.
    HTC-VLM achieves an average performance retention of \textbf{87.2\%} across seven benchmarks (GQA, VQAv2, MMBench, MME, POPE, SEED-Bench, ScienceQA-Image), outperforming the leading continuous baseline at \textbf{81.0\%} under the same one-token output budget. Attention analyses show the compressed token prioritizes the discrete anchors, supporting their role as semantic guidance. This extended study further examines token-budget behavior, cross-architecture generalization, inference efficiency, codebook and masking robustness, and formal information-theoretic properties of the hybrid bottleneck. The code is publicly available at \url{https://github.com/jushengzhang/HybridToken-VLM}.
\end{abstract}

}

\maketitle
\IEEEdisplaynontitleabstractindextext
\IEEEpeerreviewmaketitle

\section{Introduction}
\label{sec:intro}

Vision-language models (VLMs) increasingly rely on large sets of patch-level visual tokens
(\textit{e.g.}, $N{=}576$ for a single ViT image) to supply rich perceptual information to a large language model (LLM)~\cite{VIT,clip,llava,zs1,zs2,zyvit,zs3,zs4}. 
While effective, this dense coupling imposes a prohibitive quadratic attention cost 
$\mathcal{O}((N{+}L)^2)$, rapidly exhausting GPU memory and context budgets~\cite{zhuyil,zhuyil2,zhuyil3,Papa_2024}. 
A natural question thus arises: \emph{Can a VLM retain semantically useful visual information when the entire image is compressed to only a few tokens or even to a single one?}

\begin{figure}[t] \begin{center} \includegraphics[width=0.5\textwidth]{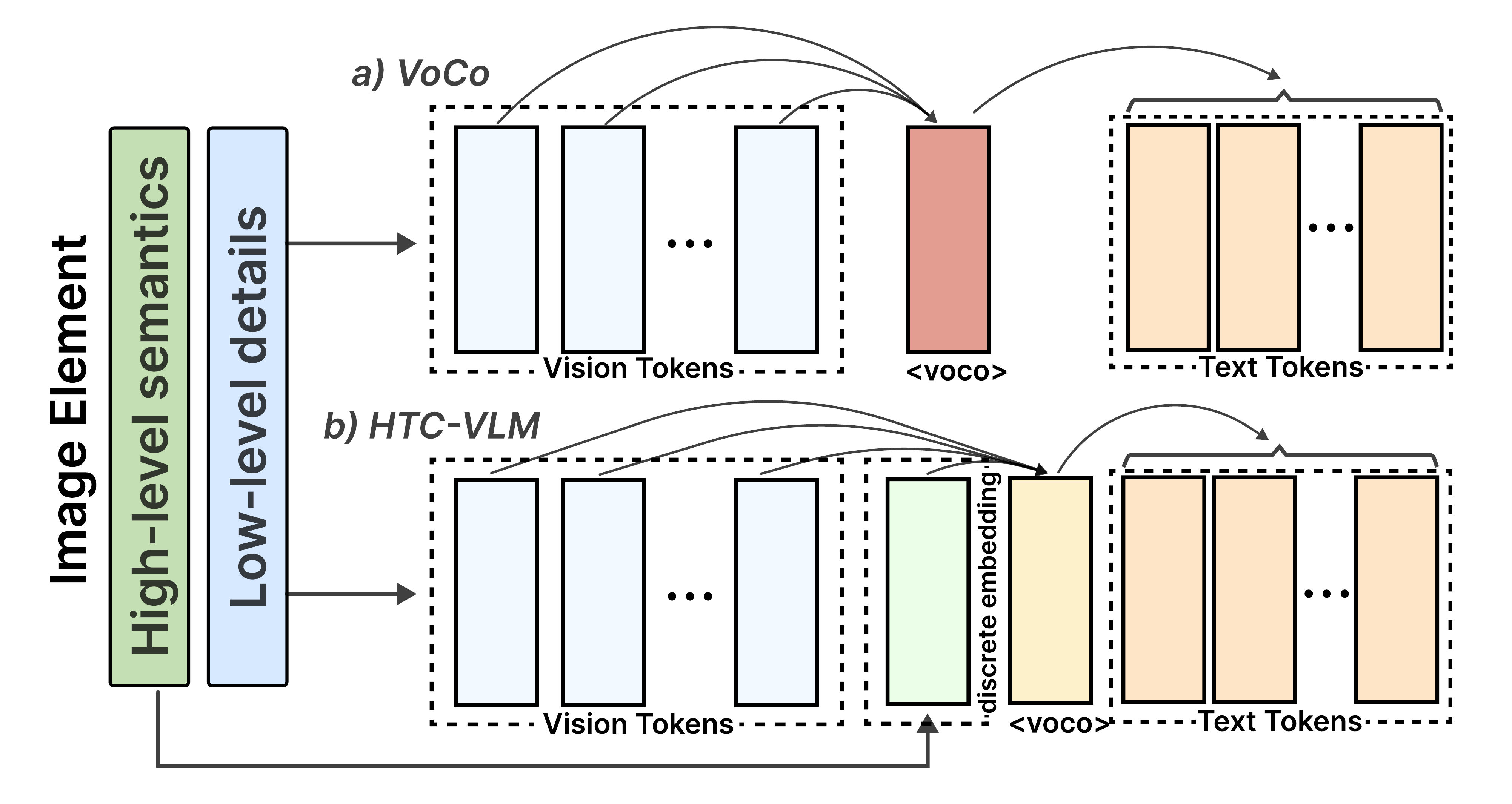} \end{center} 
\vspace{-22pt}
\caption{Vision-token compression. 
(a) VoCo-LLaMA collapses 576 patches into one \texttt{<voco>} token, losing semantic structure. 
(b) HTC-VLM adds 4 discrete semantic tokens and compresses all into one \texttt{<voco>} token, preserving semantics and visual detail.}
\label{fig:different} 
\vspace{-8pt}
\end{figure}

Existing attempts have split into two directions that exhibit complementary failure modes~\cite{zs5,yang2024lcl,kuratov-etal-2025-cramming,10030157,lstoken,9578911,yu2024imageworth32tokens,FDWR,CF}: 
\emph{Continuous compression} projects the full patch sequence into a single dense vector, reducing latency but inevitably collapsing high-level semantics as mutual information $I(v_c;S)$ drops.  
Conversely, \emph{full discretization} (\emph{e.g.}, VQ codebooks) preserves categorical semantics but discards fine-grained continuous cues (pose, texture, deformation), creating a granularity gap that limits detailed reasoning~\cite{yu2024imageworth32tokens,lisanzs,zhong2025surveyvisionlanguageactionmodelsaction}.  These behaviors are often interpreted as an unavoidable efficiency-fidelity trade-off.

We revisit this problem from a representational perspective.
By compressing all 576 ViT tokens into a \emph{single} latent, we expose a structural bottleneck: 
\emph{A one-token continuous bottleneck cannot simultaneously encode discrete semantics and continuous visual details~\cite{Papa_2024,lstoken}}.  
This motivates us to examine how much structure must be preserved \emph{before} compression.  
A key observation emerges: inserting a \emph{minimal} set of discrete semantic anchors prior to the bottleneck restores the high-level scaffolding required by downstream reasoning, while continuous tokens retain complementary fine-grained detail.
This leads to \textbf{HTC-VLM}, a hybrid compression architecture (Fig.~\ref{fig:different}) that prepends a small number of discrete semantic tokens to the continuous patch tokens and compresses them jointly into a single \texttt{<voco>} latent.  
Unlike VoCo-style methods, which compress only continuous tokens, our approach explicitly decomposes visual information into  
(i) \textbf{semantic anchors} (discrete) and  
(ii) \textbf{detail carriers} (continuous) \emph{before} fusion, following the principle that \emph{single-token compression remains expressive only if semantics and details are disentangled prior to compression}.
Empirically, HTC-VLM retains \textbf{87.2\%} of full-model performance across seven benchmarks, outperforming the best continuous-compression baseline (\textbf{81.0\%}) under the same one-token output budget.  
Attention analyses further show that the compressed latent selectively attends to the discrete anchors, validating their role as interpretable semantic carriers.

Our main contributions are summarized as follows:
\begin{itemize}[leftmargin=*, itemsep=0pt]
\item \textbf{Representational bottleneck analysis.}
We identify the expressiveness gap in single-token visual compression, formalize semantic--detail capacity conflict, and provide information-theoretic analysis explaining why purely continuous bottlenecks tend to dilute high-level semantics.
\item \textbf{Hybrid semantic and detail decomposition.}
We introduce a principled framework that injects a minimal number of discrete semantic anchors before compression, enabling disentangled fusion of discrete semantics and continuous appearance in the hybrid latent.
\item \textbf{A practical and scalable hybrid VLM architecture.}
We instantiate this principle as HTC-VLM, achieving the best average retention among the evaluated one-token baselines (87.2\%), with analyses showing that the hybrid latent helps preserve interpretable semantics and fine-grained cues.
\item \textbf{Extended empirical validation.}
Beyond the one-token setting, we study HTC-VLM across multiple token budgets and compare it with representative pruning, pooling, and sparsification baselines, showing that the hybrid design remains stable under both extreme and moderate compression regimes.
\item \textbf{Generalization, efficiency, and robustness studies.}
We further evaluate HTC-VLM on stronger VLM backbones, report inference latency and memory behavior, and ablate the discrete codebook, projector capacity, anchor selection, and attention-mask topology to clarify when and why hybrid compression remains effective.
\end{itemize}

\noindent\textbf{Extension over the Conference Version.}
This article substantially extends the conference version of HybridToken-VLM~\cite{HTC} (CVPR 2026). Following the common practice of journal extensions, we deepen the theoretical account, broaden the empirical scope, and add reproducibility and robustness analyses that were absent from the conference paper. The main extensions are summarized as follows:
\begin{itemize}[leftmargin=*, itemsep=0pt]
\item \textbf{A more complete theoretical treatment.}
The conference version mainly motivated hybrid compression empirically. This article develops a formal analysis of the semantic--detail capacity conflict, compares pruning, continuous compression, and discrete quantization under a unified information-theoretic view, and proves why the hybrid prior yields a tighter objective than a continuous-only bottleneck (Sections~\ref{sec:theory_comparison}, \ref{sec:rationale}, and \ref{sec:theory}; Table~\ref{tab:supp_comparison}).
\item \textbf{Expanded method formulation and reproducibility details.}
We provide a more explicit formulation of the continuous detail channel, the MGVQ-based discrete semantic channel, the hybrid sequence construction, the disentanglement attention mask, and the \texttt{<voco>} latent extraction. We also include full forward-pass pseudocode, mask-construction pseudocode, and detailed training hyperparameters to make the architecture easier to reproduce (Section~\ref{sec:method}, Section~\ref{sec:implementation_details}, Algorithms~\ref{alg:hybrid_forward_full}--\ref{alg:mask_full}, and Table~\ref{tab:hyperparameters}).
\item \textbf{Broader token-budget and baseline evaluation.}
Beyond the original one-token comparison, we extend the study to a wider token-budget sweep and compare HTC-VLM with representative pruning, dropping, merging, sparsification, pooling, and continuous-compression baselines. This extension clarifies how different compression paradigms degrade as the token budget moves from moderate compression to the extreme one-token regime (Section~\ref{sec:token_sweep_analysis}, Fig.~\ref{fig:token_curve}, and Table~\ref{tab:token_comparison}).
\item \textbf{Cross-architecture generalization.}
The conference version focused on the standard LLaVA-style setting. This journal version further evaluates whether the same hybrid bottleneck transfers to stronger and structurally different VLM backbones, including Qwen-VL-Chat and LLaVA-NeXT with dynamic high-resolution inputs. These results test whether the proposed semantic anchors remain useful beyond the original backbone (Section~\ref{sec:generalizability} and Table~\ref{tab:advanced_generalizability}).
\item \textbf{System-level efficiency analysis.}
We add latency, throughput, memory, and end-to-end wall-clock breakdowns to separate the benefit of reducing LLM visual-token prefill from the overhead introduced by the additional semantic encoder. This analysis makes the efficiency--accuracy trade-off explicit and shows that the hybrid branch preserves near single-token inference cost while improving accuracy over continuous-only compression (Section~\ref{sec:efficiency}, Table~\ref{tab:efficiency}, and Table~\ref{tab:latency_breakdown}).
\item \textbf{Additional robustness and component ablations.}
We include new studies on MGVQ codebook/group configuration, loss-weight sensitivity, projector capacity, semantic-anchor type, and mask topology. These experiments examine whether HTC-VLM's gain comes from the structural hybrid design, and they further justify why discrete anchors and the text-bottleneck mask are preferred (Sections~\ref{sec:codebook_ablation}--\ref{sec:robustness}, Tables~\ref{tab:mgvq_ablation}--\ref{tab:ablation_anchors_mask}).
\end{itemize}

\section{Related Work}
\label{sec:related}

\noindent\textbf{Vision-Language Models and Token Efficiency.}
Modern VLMs such as LLaVA~\cite{llava}, Qwen-VL~\cite{Qwen2.5-VL}, and GPT-4V~\cite{GPT-4o} rely on hundreds of visual tokens (\textit{e.g.}, 576 from ViT-L/14) to enable strong multimodal alignment. This dense design, however, incurs quadratic attention cost and motivates reducing the visual token budget~\cite{zhao2024msdetr}. Existing token-efficiency methods, including token merging~\cite{bolya2022token,patchdrop}, patch dropping~\cite{DynamicViT,ma2024followpose,ma2024followyouremoji,ma2025followyourmotion}, and redundancy-aware selection~\cite{liang2022patchesneedexpeditingvision,ma2025followfaster}, operate entirely in the \emph{continuous} feature space. While they effectively reduce computation, these methods degrade rapidly under extreme compression (\textit{e.g.}, $1$--$4$ tokens), where continuous features collapse and lose semantic structure.

\noindent\textbf{Continuous vs. Discrete Compression.}
A complementary direction compresses vision features after the encoder. Continuous approaches such as pooling, attention aggregation, Q-Former~\cite{instructblip}, and VoCo-LLaMA~\cite{ye2025voco} map all patches to a single dense embedding, but often suffer from \emph{semantic dilution} when diverse patches are averaged into a unimodal vector. Conversely, discrete visual tokenizers (\textit{e.g.}, VQ-VAE~\cite{lstoken}, MoVQ~\cite{zheng2022movqmodulatingquantizedvectors}, MGVQ~\cite{jia2025mgvqvqvaebeatvae}) produce compact and interpretable codes that preserve high-level semantics, but inevitably lose fine-grained appearance because quantization removes continuous variation. Neither paradigm preserves both high-level semantics and low-level details under a single-token bottleneck.

\noindent\textbf{Hybrid Representation Learning.}
Recent studies suggest that separating semantic and appearance information can improve multimodal representations, but existing approaches either require large token budgets or do not target extreme compression~\cite{lstoken,9578911,lisanzs,zs1,zs4,wang2023transhp,zhong2025surveyvisionlanguageactionmodelsaction,Papa_2024}. HTC-VLM differs by explicitly \textbf{disentangling} visual information into a discrete semantic channel and a continuous detail channel, and fusing them through a \textbf{single-token bottleneck} equipped with a disentanglement attention mask. This hybrid architecture simultaneously avoids semantic dilution and granularity loss, enabling one-token representations that remain semantically stable and detail-preserving.

\section{Problem Formulation} \label{sec:problem} 
\textbf{The Dilemma of Visual Representation.} A vision-language model (VLM) seeks to model the conditional distribution $p_\theta(Y \mid I, T)$, where an image $I$ and a textual instruction $T$ jointly generate a coherent textual response $Y$~\cite{Chen_2023,caffagni2024revolutionmultimodallargelanguage}. In architectures like LLaVA, the image is decomposed into $N=576$ patch embeddings using a pretrained vision encoder $\mathcal{E}_v$ (\textit{e.g.}, ViT-L/14) and projected into the LLM's embedding space via a trainable projector $\mathcal{P}_v$: {\small \begin{equation} V = \{ v_1, \dots, v_N \} = \mathcal{P}_v \bigl( \mathcal{E}_v(I) \bigr), \quad V \in \mathbb{R}^{576 \times d_{\text{model}}}, \label{eq:vision-embedding} \end{equation}}
\hspace{-2mm}where $d_{\text{model}} = 4096$ corresponds to the embedding dimension of modern LLMs~\cite{liu2024improved,Llama2,zs3}. This representation facilitates multimodal alignment by mapping visual features into a shared semantic space, but its high dimensionality introduces profound computational and informational challenges, as detailed below. \subsection{The Scaling Imperative and Compression Objective} \label{subsec:scaling-objective} The LLM's self-attention mechanism~\cite{li2023blip,alayrac2022flamingovisuallanguagemodel}, defined as $A = \text{softmax}\bigl(\frac{Q K^T}{\sqrt{d_k}}\bigr) V$, scales quadratically with the total sequence length, $\mathcal{O}((N + L)^2)$, where $L$ is the text token count and $d_k$ is the attention head dimension (typically $d_k = 64$). For $N=576$ patch embeddings, this results in a per-layer memory complexity of $\Theta(N^2 d_{\text{model}})$, approximately $1.07 \times 10^9$ floating-point operations for $d_{\text{model}} = 4096$ on a single layer, rapidly exhausting GPU memory (\textit{e.g.}, 24GB VRAM) and saturating context windows (\textit{e.g.}, 4096 tokens). This quadratic bottleneck, exacerbated by multi-head attention across $h$ heads, motivates compressing $V$ to a single token, reducing the visual term to $\mathcal{O}(L^2)$ and yielding a theoretical speedup of $576^2 \approx 3.3 \times 10^5$ in attention computations. The optimal compressor $\mathcal{C}$ is formulated as: {\small \begin{equation} \mathcal{C}^\star = \arg\min_{\mathcal{C}} \mathbb{E}_{(I,T,Y) \sim \mathcal{D}} \left[ \mathcal{L} \bigl( Y, \mathrm{VLM} \bigl( T, \mathcal{C} \bigl( \mathcal{E}_v(I) \bigr) \bigr) \bigr) \right], \label{eq:compression-objective} \end{equation}}
\hspace{-3mm}where $\mathcal{L}$ is typically the cross-entropy loss $\mathcal{L} = -\sum_{y \in Y} \log p_\theta(y \mid T, V_c)$, and $\mathcal{D}$ is the joint distribution over $(I, T, Y)$. This optimization problem, however, reveals a critical trade-off: compressing to $|V_c| = 1$ risks diminishing the information content $I(V_c; Y)$, necessitating a balance between computational efficiency and representational fidelity, as explored next. 

\subsection{The Representation Dilemma} \label{subsec:dilemma} The compression challenge manifests in two paradigms, each with distinct limitations. \emph{Continuous compression} transforms $V$ into a single vector $v_c = \mathcal{C}_{\text{cont}}(V) \in \mathbb{R}^{d_{\text{model}}}$, often via global pooling or attention aggregation. Information-theoretically, this process reduces the entropy $H(V)$ to $H(v_c)$, where the mutual information $I(v_c; S)$ with high-level semantics $S$ (\textit{e.g.}, object identities, spatial relations) diminishes. This \emph{semantic dilution} arises because averaging convolves diverse patch distributions into a unimodal representation, lowering $H(v_c)$ below the threshold needed for disambiguation. For instance, averaging patches of a ``dog'' and a ``cat'' yields $v_c$ with insufficient entropy to distinguish species, forcing the LLM to rely on ambiguous prior distributions, \textit{i.e.}, leading to errors in tasks like object classification (see Section \ref{subsec:method-continuous} for our approach to mitigate this). In contrast, \emph{discrete compression} via vector quantization maps $I$ to indices $k = \arg\min_j \| f(I) - c_j \|_2^2$, where $\{c_j\}_{j=1}^K$ is a codebook of size $K$. This preserves interpretability by clustering semantic modes, but introduces a \emph{granularity gap} due to quantization noise $\epsilon = f(I) - c_k$. The mutual information $I(k; D)$ with low-level details $D$ (\textit{e.g.}, texture gradients, pose angles) is reduced, as continuous feature variance is discretized into discrete bins. Practically, this manifests when a Golden Retriever on grass and a Poodle on sand map to the same $k$, erasing contextual cues critical for fine-grained tasks like pose estimation or texture recognition (see Section \ref{subsec:method-discrete} for our hybrid resolution). Neither paradigm optimizes the joint information $I(V_c; S, D)$ while minimizing redundancy $I(S; D \mid V_c)$, highlighting the need for a disentangled representation. We formalize this semantic--detail capacity conflict and prove why the hybrid prior yields a tighter objective in Section~\ref{sec:theory}.

\subsection{A Guiding Question for Disentangled Compression} \label{subsec:guiding-question} The trade-off between semantic dilution and granularity gap suggests that a compact representation must disentangle $S$ and $D$ to maximize their joint contribution $I(V_c; S) + I(V_c; D)$ while reducing conditional dependence $I(S; D \mid V_c)$. This requires a representation where the compressed $V_c$ acts as a sufficient statistic for both $S$ and $D$, satisfying the Markov condition $S \perp D \mid V_c$. This insight frames our central inquiry: \begin{tcolorbox}[colback=blue!5!white, colframe=blue!60!black, boxrule=0.5pt, arc=3pt, left=6pt, right=6pt, top=2pt, bottom=2pt] \emph{How can we craft an ultra-compact visual representation that \textbf{disentangles} high-level, discrete semantics from low-level, continuous appearance, thereby escaping both semantic dilution and the granularity gap?} \end{tcolorbox} This demands a hybrid framework where complementary channels encode orthogonal information, preserving diversity across $S$ and $D$. HTC-VLM, introduced in Section~\ref{sec:method}, proposes a dual-channel architecture with a theoretically grounded bottleneck to achieve this disentanglement, as detailed below.

\section{Method} \label{sec:method} 

\begin{figure*}[t] \begin{center} \includegraphics[width=1 \textwidth]{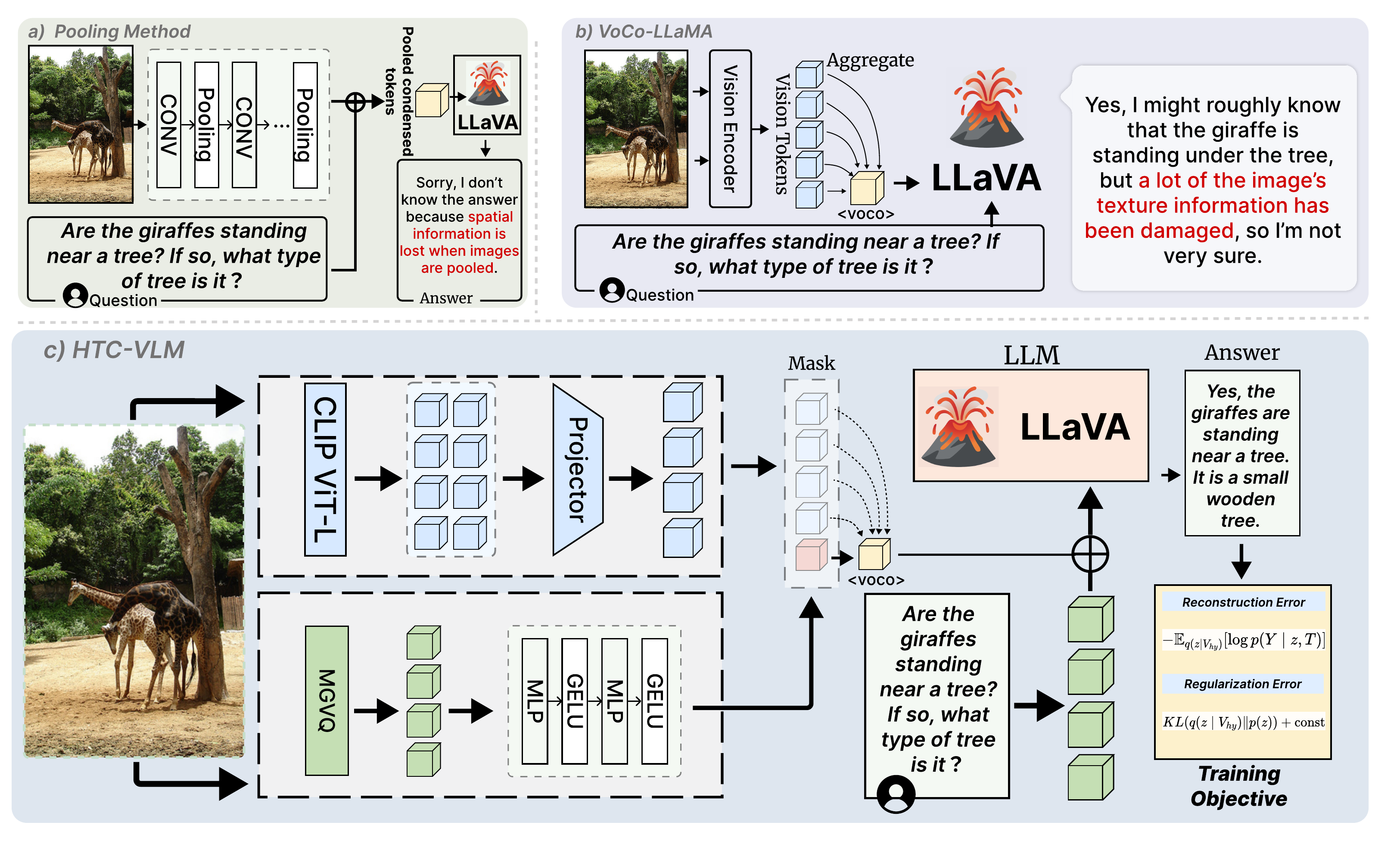} \end{center} 
\vspace{-20pt}
\caption{Comparison of visual token compression strategies. (a) \textbf{Pooling Method:} visual embeddings are averaged or pooled before being fused with text inputs. (b) \textbf{VoCo-LLaMA:} compresses 576 visual tokens into a single \texttt{<voco>} token. (c) \textbf{HTC-VLM (ours):} introduces a hybrid representation with a continuous channel ($D$) encoding 576 patch embeddings and a discrete channel ($S$) generating 4 semantic tokens via MGVQ. The hybrid sequence $[v_d; V]$ is compressed into a trainable \texttt{<voco>} token under the disentanglement mask $M_{hy}$, producing latent $z$ that preserves both semantics and fine-grained details.} \label{fig:main_flow}
\vspace{-4pt}
 \end{figure*} 
As illustrated in Fig.~\ref{fig:main_flow}, HTC-VLM tackles the guiding question from Section \ref{subsec:guiding-question} by disentangling high-level semantics $S$ (\textit{e.g.}, object categories, spatial layouts) and low-level details $D$ (\textit{e.g.}, textures, poses) into distinct channels, fused through a disentanglement bottleneck that compresses the representation into a single token. This design emerged from the dilemma in Section \ref{subsec:dilemma}: initial experiments compressing $V$ (Eq. \ref{eq:vision-embedding}) to a single continuous token revealed a collapse in mutual information $I(v_c; S) \to 0$ due to variance loss, as the entropy $H(v_c) \ll H(V)$ failed to capture semantic diversity. Iterative trials with discrete augmentations, guided by information-theoretic metrics, identified that four discrete semantic tokens from a vector quantizer (VQ) restored $I(v_d; S)$, inspiring HTC-VLM's architecture~\cite{9578911,tishby2000informationbottleneckmethod}. This semantic--detail capacity conflict and the hybrid resolution are formalized in Section~\ref{sec:theory_comparison}, while Section~\ref{sec:rationale} provides the Information Bottleneck and variational rationale for the hybrid design and masking topology. \begin{tcolorbox}[colback=blue!5!white, colframe=blue!60!black, boxrule=0.5pt, arc=3pt, left=6pt, right=6pt, top=2pt, bottom=2pt] \textbf{Core Idea.} HTC-VLM disentangles visual information into a continuous channel for $D$ and a discrete channel for $S$, fused via a disentanglement bottleneck to optimize the joint information $I(V_c; S, D)$ while minimizing redundancy $I(S; D \mid V_c)$. \end{tcolorbox} 

\subsection{Exploratory Decomposition and Channel Design} \label{subsec:method-exploration} Our development began with a compression experiment on $V$, where reducing it to one token via averaging or attention pooling diminished $H(V)$ to $H(v_c)$, losing semantic structure as $I(v_c; S) \approx 0$ (Section \ref{subsec:dilemma}). To address this, we explored discrete representations, evaluating multiple VQ models. MGVQ~\cite{jia2025mgvq}, with its multi-group quantization (8 groups, 16384 codebook size, 16x downsampling), emerged as optimal due to its ability to cluster diverse semantic modes, decomposing $I$ into dual channels that maximize $I(V; D) + I(q; S)$. Section~\ref{sec:codebook_ablation} further ablates the MGVQ codebook size and group number, supporting this default configuration as a stable trade-off between semantic capacity and optimization robustness.
\subsubsection{Continuous Channel: Encoding Low-Level Details \texorpdfstring{$D$}{D}} \label{subsec:method-continuous} To preserve the information content $I(V; D)$ and counter the granularity gap (Section \ref{subsec:dilemma}), we employ a pretrained vision encoder $\mathcal{E}_v$ (CLIP ViT-L/14) and a trainable linear projector $\mathcal{P}_v$ to generate a sequence of $N=576$ patch embeddings: \begin{equation} V = \{ v_i \}_{i=1}^{576} = \mathcal{P}_v \bigl( \mathcal{E}_v(I) \bigr), \quad v_i \in \mathbb{R}^{4096}, \label{eq:continuous-channel} \end{equation} where $\mathcal{P}_v: \mathbb{R}^{d_{\text{vision}}} \to \mathbb{R}^{4096}$ is a learned linear transformation aligning patch features with the LLM's embedding space. This high-dimensional manifold captures fine-grained details like texture gradients and pose variations, ensuring $V$ retains a rich representation of $D$ with entropy $H(V) \propto \log |\mathcal{M}_D|$, where $\mathcal{M}_D$ is the detail manifold. 
\subsubsection{Discrete Channel: Encoding High-Level Semantics \texorpdfstring{$S$}{S}} \label{subsec:method-discrete} To restore $I(q; S)$ and mitigate semantic dilution (Section \ref{subsec:dilemma}), we leverage MGVQ to quantize $I$ into a feature vector $q \in \mathbb{R}^{14112}$, reflecting its multi-group structure. This is projected to a sequence of discrete anchor tokens $v_d \in \mathbb{R}^{4 \times 4096}$ via a two-layer MLP $\mathcal{P}_d$ with GELU activations: \begin{equation} \resizebox{\columnwidth}{!}{$ q = \mathcal{Q}(I), \quad v_d = \mathcal{P}_d(q) = \text{GELU} \bigl( W_2 \cdot \text{GELU} \bigl( W_1 \cdot q \bigr) \bigr), $} \label{eq:discrete-channel} \end{equation} where $W_1 \in \mathbb{R}^{16384 \times 14112}$ and $W_2 \in \mathbb{R}^{16384 \times 16384}$ are weight matrices, and $\text{GELU}(x) = x \Phi(x)$ (where $\Phi$ is the Gaussian CDF) introduces non-linearity. MGVQ's quantization minimizes the reconstruction error $\mathbb{E} \| I - \mathcal{Q}^{-1}(q) \|_2^2$, clustering $S$ into discrete modes (\textit{e.g}., ``dog on grass''), with $v_d$ serving as four low-dimensional anchors that preserves $I(v_d; S) \approx H(S)$ under codebook constraints. 

\subsection{Fusion and Disentanglement Bottleneck: Theoretical Framework} \label{subsec:method-fusion} The channels are fused by prepending $v_d$ to $V$, forming a hybrid sequence: \begin{equation} V_{hy} = [v_d; V] \in \mathbb{R}^{580 \times 4096}, \label{eq:hybrid-representation} \end{equation} followed by a trainable \texttt{<voco>} token. The Disentanglement Attention Mask $M_{hy}$ is defined over the full input $X = [V_{hy}; \texttt{<voco>}; W]$, where $W$ are text embeddings: 
\begin{equation}
M_{hy}(i, j) =
\begin{cases}
0, & \text{if } i \in W \text{ and } j \in \texttt{\textless voco\textgreater}, \\
-\infty, & \text{if } i \in W \text{ and } j \in V_{hy}, \\
0, & \text{otherwise}.
\end{cases}
\label{eq:attention-mask}
\end{equation}
This mask ensures text attends only to \texttt{<voco>}, while \texttt{<voco>} integrates $V_{hy}$. Theoretically, this bottleneck approximates a variational autoencoder (VAE), where \texttt{<voco>} represents a latent variable $z$ with posterior $p(z \mid V_{hy})$. The objective is to minimize the Kullback-Leibler divergence $KL(p(V_{hy} \mid z) \| p(V_{hy}))$, guided by $v_d$ to disentangle $S$ and $D$. The evidence lower bound (ELBO) for this process is: \begin{equation} \resizebox{\columnwidth}{!}{$ \log p(Y \mid T, I) \geq \mathbb{E}_{q(z \mid V_{hy})} \left[ \log p(Y \mid z, T) \right] - KL(q(z \mid V_{hy}) \| p(z)), $} \label{eq:elbo} \end{equation} where $q(z \mid V_{hy}) = p(\texttt{<voco>} \mid V_{hy}; \theta)$ is learned, and $p(z)$ is a prior (\textit{e.g.}, $\mathcal{N}(0, I)$). The compression ratio of 580-to-1 is achieved by optimizing $z$ to maximize the mutual information: \begin{equation} \mathcal{I}(z; V_{hy}) = \mathbb{E}_{p(V_{hy})} \left[ \log \frac{p(V_{hy}, z)}{p(V_{hy}) p(z)} \right], \label{eq:mutual-information} \end{equation} where $v_d$ biases $z$ toward $S$, and $V$ contributes $D$, reducing $I(S; D \mid z)$ via $M_{hy}$'s constraint. \begin{tcolorbox}[colback=blue!5!white, colframe=blue!60!black, boxrule=0.5pt, arc=3pt, left=6pt, right=6pt, top=2pt, bottom=2pt] \textbf{Disentanglement Bottleneck.} The \texttt{<voco>} token compresses $V_{hy}$ into a latent $z$, with $v_d$ enforcing disentangled encoding of $S$ and $D$ via $M_{hy}$, optimizing $\mathcal{I}(z; S) + \mathcal{I}(z; D)$. \end{tcolorbox}

\subsection{Training Objective and Optimization Dynamics} \label{subsec:method-training} The training objective is the expected autoregressive loss: \begin{equation} \resizebox{\columnwidth}{!}{$ \mathcal{L}_{\text{HTC}} = -\mathbb{E}_{p(I,T,Y)} \left[ \sum_{i=1}^{|Y|} \log p_\theta(y_i \mid y_{<i}, \texttt{<voco>}, T; M_{hy}) \right], $} \label{eq:training-objective} \end{equation} where $M_{hy}$ shapes the gradient flow. This loss can be decomposed into a variational lower bound, aligning with the ELBO (Eq. \ref{eq:elbo}): \begin{equation} \resizebox{\columnwidth}{!}{$ \mathcal{L}_{\text{HTC}} \approx -\mathbb{E}_{q(z \mid V_{hy})} \left[ \log p(Y \mid z, T) \right] + KL(q(z \mid V_{hy}) \| p(z)) + \text{const}, $} \label{eq:decomposed-loss} \end{equation} where the first term is the reconstruction error, and the second term regularizes $z$. The mask $M_{hy}$ constrains $q(z \mid V_{hy})$ to depend on \texttt{<voco>}, with $v_d$ acting as a prior anchor for $S$. Gradient dynamics reveal that $\frac{\partial \mathcal{L}}{\partial v_d}$ enhances semantic clustering (\textit{e.g.}, maximizing $I(v_d; S)$), while $\frac{\partial \mathcal{L}}{\partial V}$ refines $D$'s variance, achieving a disentangled latent space. This optimization leverages the bottleneck to outperform single-channel baselines by preserving $I(\texttt{<voco>}; S, D)$, as validated in subsequent experiments. \begin{tcolorbox}[colback=blue!5!white, colframe=blue!60!black, boxrule=0.5pt, arc=3pt, left=6pt, right=6pt, top=2pt, bottom=2pt] \textbf{Training Pressure Redirected.} The mask $M_{hy}$ re-routes gradients to enforce a disentangled latent $z$, optimizing $I(\texttt{<voco>}; S, D)$ via variational inference. \end{tcolorbox} Algorithm~\ref{alg:framework} summarizes this high-level HTC-VLM forward pass. \begin{algorithm}[!t] \caption{HTC-VLM Forward Pass} \label{alg:framework} \begin{algorithmic}[1] \Require Image $I$, Text $T$, components ($\mathcal{E}_v$, $\mathcal{P}_v$, $\mathcal{Q}$, $\mathcal{P}_d$, $\mathcal{E}_t$, $\mathcal{L}_{\mathrm{LLM}}$) \State $V \gets \mathcal{P}_v(\mathcal{E}_v(I))$ \Comment{Generate 576 patch embeddings} \State $q \gets \mathcal{Q}(I)$ \Comment{MGVQ quantizes to 14112 features} \State $v_d \gets \mathcal{P}_d(q)$ \Comment{MLP projection to discrete embedding} \State $V_{hy} \gets [v_d; V]$ \Comment{Construct 580-token hybrid} \State $W \gets \mathcal{E}_t(T)$ \Comment{Encode text} \State $X \gets [V_{hy}; \texttt{<voco>}; W]$ \Comment{Integrate with \texttt{<voco>} token} \State $M_{hy} \gets \text{CreateDisentanglementAttentionMask}(X)$ \Comment{Apply disentanglement mask} \State \text{Logits} $\gets \mathcal{L}_{\mathrm{LLM}}(X, M_{hy})$ \Comment{Compute logits} \State \Return \text{Logits} \end{algorithmic}
 \end{algorithm}

\vspace{-3mm}
\section{Theoretical Comparison and Complexity Analysis}
\label{sec:theory_comparison}

In this section, we provide a structured comparison of HTC-VLM against existing visual compression paradigms. We analyze the methods from three perspectives: \textit{Representational Manifold}, \textit{Inference Complexity} (specifically the attention bottleneck), and \textit{Information-Theoretic Bounds}. This comparison contextualizes the formal proofs in Section~\ref{sec:theory}.

Table \ref{tab:supp_comparison} summarizes the fundamental limitations of previous approaches and formally demonstrates how HTC-VLM overcomes the \textit{Semantic-Detail Capacity Conflict} (Theorem 1) to achieve a tighter Evidence Lower Bound (Theorem 2).

\begin{table*}[!t]
\centering
\small
\caption{\textbf{Formal Comparison of Visual Compression Paradigms.} We contrast HTC-VLM with Pruning, Continuous Compression, and Discrete Quantization. \textbf{Notation:} $N$ is the original patch count (576), $M$ is the pruned count ($M<N$), $d$ is the embedding dimension, $I(\cdot;\cdot)$ denotes mutual information, and $S, D$ represent Semantics and Details, respectively.}
\vspace{-3mm}
\label{tab:supp_comparison}
\renewcommand{\arraystretch}{1.5} 
\setlength{\tabcolsep}{3pt}
\begin{tabularx}{\textwidth}{@{}p{0.16\textwidth}p{0.16\textwidth}lYp{0.16\textwidth}@{}}
\toprule
\textbf{\makecell[l]{Method \\ Paradigm}} & \textbf{\makecell[l]{Visual \\ Representation}} & \textbf{\makecell[l]{Attention \\ Complexity}} & \textbf{\makecell[l]{Information-Theoretic \\ Limitation}} & \textbf{\makecell[l]{Proof / \\ Theorem Ref.}} \\
\midrule

\textbf{Structured Pruning}\newline \textit{(\textit{e.g.}, ToMe, FastV)} & 
Subset of Patches\newline $V_{sub} \subset \mathbb{R}^{M \times d}$ & 
$\mathcal{O}(M^2)$ & 
\textbf{Structure Collapse}: Direct removal of tokens breaks topological priors. No explicit maximization of $I(V_{sub}; S)$ or $I(V_{sub}; D)$. Non-smooth degradation. & 
Empirical Observation\newline (Section~\ref{sec:token_sweep_analysis}, Fig.~\ref{fig:token_curve}) \\
\midrule

\textbf{Continuous Compression}\newline \textit{(\textit{e.g.}, VoCo-LLaMA)} & 
Single Vector\newline $v_c \in \mathbb{R}^{1 \times d}$ & 
$\mathbf{\mathcal{O}(1)}$ & 
\textbf{Capacity Conflict}: Bounded entropy $H(v_c)$ cannot accommodate distinct modes of $S$ and variance of $D$. \textit{Result:} $I(v_c; S) \to 0$ as $I(v_c; D)$ increases. & 
\textbf{Theorem 1}\newline (Section~\ref{subsec:proof_theorem1}) \\
\midrule

\textbf{Discrete Quantization}\newline \textit{(\textit{e.g.}, VQ-based)} & 
Discrete Codes\newline $z_d \in \mathbb{Z}^{k}$ & 
$\mathbf{\mathcal{O}(1)}$ & 
\textbf{Granularity Gap}: Information about details $D$ is strictly upper-bounded by codebook size $|\mathcal{C}|$. \textit{Result:} $I(z_d; D) \le \log |\mathcal{C}|$. & 
Info. Theory Axiom\newline (Section~\ref{sec:rationale}) \\
\midrule

\textbf{HTC-VLM (Ours)} & 
\textbf{Hybrid Latent}\newline $z \sim q(z|v_d, V)$ & 
$\mathbf{\mathcal{O}(1)}$ & 
\textbf{Disentangled Sufficiency}: The hybrid prior $v_d$ lowers $H(S|z)$, allowing $z$ to dedicate capacity to $D$. \textit{Result:} Maximize joint info $I(z; S) + I(z; D)$. & 
\textbf{Theorem 2}\newline (Section~\ref{subsec:proof_theorem2}) \\
\bottomrule
\end{tabularx}
\vspace{-2mm}
\end{table*}

\subsection{Detailed Comparative Analysis}
\label{subsec:detailed_analysis}

The comparison in Table~\ref{tab:supp_comparison} highlights three critical insights regarding the efficiency-fidelity trade-off in visual compression:

\noindent\textbf{1. The ``Pseudo-Efficiency'' of Structured Pruning.}
While structured pruning methods (\textit{e.g.}, ToMe, FastV) reduce the token count from $N$ to $M$, they fail to eliminate the fundamental quadratic bottleneck. Since attention complexity scales as $\mathcal{O}(M^2)$, even a moderate reduction (\textit{e.g.}, $M=64$) retains significant computational cost compared to single-token approaches ($\mathcal{O}(1)$). Furthermore, Fig.~\ref{fig:token_curve} demonstrates that pruning suffers from \textit{structure collapse}: removing tokens without a semantic guide destroys the topological coherence required for spatial reasoning, leading to sharp performance degradation in the low-token regime ($M < 32$).

\noindent\textbf{2. Entropy Domination in Continuous Compression.}
Table~\ref{tab:supp_comparison} formalizes the failure mode of continuous-only methods (\textit{e.g.}, VoCo-LLaMA) via \textbf{Theorem 1}. In a single continuous vector $v_c$, the high-frequency variance of visual details (texture, noise) typically possesses much higher entropy than discrete semantic categories ($H(D) \gg H(S)$). Under standard reconstruction objectives, the limited capacity of $v_c$ becomes saturated by these high-entropy details. This phenomenon, which we term \textit{Entropy Domination}, forces the mutual information with semantics $I(v_c; S)$ to vanish, resulting in the ``semantic dilution'' observed in our qualitative heatmaps.

\noindent\textbf{3. The Hybrid Resolution.}
HTC-VLM is the only paradigm that simultaneously achieves $\mathcal{O}(1)$ complexity and high information fidelity. By offloading the semantic burden to discrete anchors $v_d$, the hybrid architecture effectively bypasses the capacity conflict. 
As proved in \textbf{Theorem 2}, the hybrid latent $z$ does not need to relearn the semantic categories from scratch; instead, it acts as a \textit{residual} encoder that conditions on $v_d$ to fill in the fine-grained details $D$. This factorization allows HTC-VLM to maintain the structural stability of discrete methods while retaining the textural expressiveness of continuous methods, yielding a strictly tighter Evidence Lower Bound (ELBO) for the generative task.

\subsection{Rationale}
\label{sec:rationale}

In this section, we provide a deeper theoretical justification for the architectural choices in HTC-VLM, specifically the necessity of the hybrid representation and the design of the disentanglement bottleneck. We ground our analysis in the Information Bottleneck (IB) principle and variational inference.

\subsubsection{Theoretical Justification for Hybrid Representation}
\label{subsec:theory_hybrid}

The core challenge of single-token visual compression is optimizing the trade-off between compression rate and information retention. Formally, let $I$ be the input image, $Y$ be the target text response, and $Z$ be the compressed latent representation (the $\texttt{<voco>}$ token). Following the Information Bottleneck principle, our goal is to maximize the mutual information $I(Z; Y)$ while minimizing the complexity of $Z$, often approximated by minimizing $I(Z; I)$ or constraining the dimensionality $|Z|$.

Standard continuous compression (\textit{e.g.}, global pooling) maps the high-dimensional manifold of $I$ to a single vector $z_c$. While this retains continuous variance, it suffers from \textit{semantic dilution}. The entropy $H(z_c)$ is dominated by high-frequency noise (texture, lighting), making the extraction of categorical semantics $S$ difficult:
\begin{equation}
    I(z_c; S) \approx H(S) - H(S|z_c) \rightarrow 0,
\end{equation}
as the conditional entropy $H(S|z_c)$ remains high due to the ``averaging'' of distinct semantic features.

Conversely, purely discrete quantization (\textit{e.g.}, VQ) maps $I$ to a discrete code $z_d \in \mathbb{Z}$. While this maximizes $I(z_d; S)$ by collapsing intra-class variance, it introduces an irreducible \textit{granularity gap} for low-level details $D$:
\begin{equation}
    I(z_d; D) \leq \log |\mathcal{C}|,
\end{equation}
where $|\mathcal{C}|$ is the codebook size. The information about fine-grained appearance is strictly upper-bounded by the quantization resolution.

\textbf{Our Solution:} HTC-VLM explicitly decomposes the information source into two channels: discrete anchors $v_d$ maximizing $I(v_d; S)$ and continuous patches $V$ preserving $I(V; D)$. The hybrid fusion ensures the compressed latent $Z_{\texttt{voco}}$ acts as a sufficient statistic for the joint distribution:
\begin{equation}
    I(Z_{\texttt{voco}}; Y) \approx I(Z_{\texttt{voco}}; v_d) + I(Z_{\texttt{voco}}; V) - \mathcal{R},
\end{equation}
where $\mathcal{R}$ represents redundancy. By providing $v_d$ as a ``semantic prior,'' we lower the entropy barrier for the model to capture $S$, allowing the continuous capacity of $Z_{\texttt{voco}}$ to be dedicated to encoding details $D$.

\subsubsection{Design of the Disentanglement Attention Mask}
\label{subsec:rationale_mask}

The default mask $M_{\mathrm{hy}}$ (Eq.~(6)) implements a \textbf{text--vision bottleneck}, not a blanket ban on visual self-attention. Language tokens $W$ are forbidden from attending to $V_{\mathrm{hy}}$ and must read the compressed $\texttt{<voco>}$ latent $z$, so $p(Y \mid T, I)$ cannot bypass the single-token visual carrier. All other positions remain unmasked: $V_{\mathrm{hy}}$ tokens (discrete anchors and patches) may attend to one another, and $\texttt{<voco>}$ may attend to the full hybrid sequence, preserving local structure before aggregation.

This design aligns with the variational view in Section~\ref{subsec:method-training}: $q(z \mid V_{\mathrm{hy}})$ is learned through $\texttt{<voco>}$, while the decoder's text stream is forced to condition on $z$ rather than on 580 raw visual keys. Intra-$V_{\mathrm{hy}}$ mixing supplies contextualized patch features for $\texttt{<voco>}$ to pool; the bottleneck constraint applies specifically to the \emph{text} queries.

We also ablate a stricter \textbf{star-graph} variant that additionally blocks $V_{\mathrm{hy}} \leftrightarrow V_{\mathrm{hy}}$ (Section~\ref{sec:robustness}). That variant can over-restrict visual mixing before compression; our default mask keeps intra-visual attention while still enforcing the text-only path to $z$.

\subsubsection{Pre-fusion vs. Post-fusion Strategy}
\label{subsec:rationale_fusion}

Our ablation studies confirm that placing discrete tokens \textit{before} continuous patches (Pre-fusion) yields superior performance. We attribute this to the ``Prompting Effect'' in autoregressive transformers. 

The discrete tokens $v_d$ serve as high-level \textit{meta-instructions} or \textit{schema} for the image. By processing them first, the attention mechanism establishes a semantic context (\textit{e.g.}, ``there is a dog and a tree'') \textit{before} processing the 576 noisy detail patches. Mathematically, this conditions the attention weights $\alpha_{ij}$ of the detail tokens on the semantic anchors:
\begin{equation}
    \alpha_{ij} = \mathrm{softmax}\left( \frac{Q(v_j) K([v_d; V])^T}{\sqrt{d}} \right).
\end{equation}
When $v_d$ occupies the prefix positions, it dominates the attention distribution's initial focus, effectively guiding the query optimization for the subsequent compression step.

\subsection{Formal Proofs and Information-Theoretic Analysis}
\label{sec:theory}

In this section, we provide formal proofs for the information-theoretic claims developed in this paper. We model the visual compression problem using Information Bottleneck (IB) theory to demonstrate why single-token continuous compression suffers from \textit{Entropy Domination} and how our hybrid approach resolves this capacity conflict.

\begin{table}[!t]
    \centering
    \small
    \caption{\textbf{Notation for Information-Theoretic Analysis.}}
    \label{tab:notation_theory}
    \vspace{-3mm}
    \renewcommand{\arraystretch}{1.12}
    \setlength{\tabcolsep}{4pt}
    \begin{tabularx}{\columnwidth}{@{}
        P{0.22\columnwidth}
        L{0.72\columnwidth}
    @{}}
        \toprule
        \textbf{Symbol} & \textbf{Definition} \\
        \midrule
        $I$ & Input Image \\
        $V$ & Visual features, \textit{i.e.}, a sequence of patch embeddings \\
        $S$ & High-level semantics, \textit{e.g.}, object class; discrete and low entropy \\
        $D$ & Low-level details, \textit{e.g.}, texture or noise; continuous and high entropy \\
        $z_c$ & Compressed latent in Continuous-Only methods, \textit{e.g.}, VoCo \\
        $z$ & Compressed continuous latent in HTC-VLM \\
        $v_d$ & Discrete semantic anchors in HTC-VLM \\
        $C$ & Channel capacity of a single-token bottleneck \\
        $H(\cdot)$ & Shannon entropy \\
        $I(\cdot; \cdot)$ & Mutual information \\
        \bottomrule
    \end{tabularx}
    \vspace{-3mm}
\end{table}

\subsubsection{Preliminaries and Notation}
We define the relevant random variables and notations in Table~\ref{tab:notation_theory}.

\subsubsection{Theorem 1: Entropy Domination}
\label{subsec:proof_theorem1}

\textbf{Theorem 1 (Entropy Domination in Continuous Compression).} 
\textit{Given a single-token continuous bottleneck $z_c \in \mathbb{R}^d$ with limited capacity $C$, if the visual source $V$ contains both semantics $S$ and details $D$ where $H(D) \gg H(S)$, and the training objective prioritizes reconstruction $\mathcal{L}_{rec}$, then the mutual information $I(z_c; S)$ is bounded by:}
\begin{equation}
    I(z_c; S) \le C - I(z_c; D)
\end{equation}
\textit{As the model maximizes detail retention $I(z_c; D)$ to lower reconstruction error, $I(z_c; S)$ approaches 0.}

\noindent\textbf{Proof.}
Let the capacity of the bottleneck be upper-bounded by $C \approx H(z_c)$. The goal of the compressor is to capture information about the source $V$. Assuming $V$ is a composition of Semantics $S$ and Details $D$, we have:
\begin{equation}
    I(z_c; V) = I(z_c; S, D)
\end{equation}
Using the Chain Rule for Mutual Information:
\begin{equation}
    I(z_c; S, D) = I(z_c; D) + I(z_c; S | D)
\end{equation}
Assuming $S$ and $D$ are independent sources (disentangled in nature, though entangled in pixel space), $I(z_c; S|D) \approx I(z_c; S)$. The bottleneck constraint implies:
\begin{equation}
    I(z_c; D) + I(z_c; S) \le H(z_c) \le C
\end{equation}
Rearranging for semantic information:
\begin{equation}
    I(z_c; S) \le C - I(z_c; D)
\end{equation}
In natural images, the entropy of pixel-level variance (noise, texture $D$) is orders of magnitude higher than object-level semantics ($S$). Standard reconstruction losses (\textit{e.g.}, MSE on features or Cross-Entropy on pixels) force the model to capture high-frequency components to minimize the loss term. Thus, the optimization drives $I(z_c; D) \to C$.
Consequently, the capacity remaining for semantics vanishes:
\begin{equation}
    \lim_{I(z_c; D) \to C} I(z_c; S) = 0
\end{equation}
This confirms the \textit{Semantic Dilution} phenomenon observed in VoCo-LLaMA. \hfill $\square$

\subsubsection{Theorem 2: Tighter ELBO with Hybrid Prior}
\label{subsec:proof_theorem2}

\textbf{Theorem 2 (Hybrid Superiority).} 
\textit{The HTC-VLM hybrid objective provides a strictly tighter Evidence Lower Bound (ELBO) than the continuous-only baseline by conditioning the posterior on discrete semantic anchors.}

\noindent\textbf{Proof.}
Consider the standard Variational Autoencoder (VAE) objective used in continuous compression (like VoCo), which maximizes the ELBO:
\begin{equation}
    \label{eq:standard_elbo}
    \mathcal{L}_{\text{cont}} = \mathbb{E}_{q(z|V)}[\log p(Y|z, T)] - D_{KL}(q(z|V) || p(z))
\end{equation}
where $p(z) = \mathcal{N}(0, I)$ is an uninformative prior. The latent $z$ must encode both $S$ and $D$.

In HTC-VLM, we introduce a discrete variable $v_d$ (the anchors) which is extracted deterministically or stochastically from $V$. The generation process is now conditioned on both the compressed latent $z$ and the anchors $v_d$. The hybrid ELBO is:
{\small
\begin{equation}
    \label{eq:hybrid_elbo}
    \mathcal{L}_{\text{hybrid}} = \mathbb{E}_{q(z|V, v_d)}[\log p(Y|z, v_d, T)] - D_{KL}(q(z|V, v_d) || p(z|v_d))
\end{equation}}
Here, $p(z|v_d)$ is a \textit{conditional prior}. Since $v_d$ captures the high-level semantic modes (\textit{e.g.}, ``dog''), the prior $p(z|v_d)$ is centered around the semantic mean, having significantly lower variance than the standard prior $p(z)$.

\textbf{Step 1: Reconstruction Gain.} 
Conditioning on $v_d$ simplifies the task for $z$. The latent $z$ only needs to encode the \textit{residual} details (Details $D$) relative to the semantic anchors. Thus:
\begin{equation}
    \mathbb{E}[\log p(Y|z, v_d)] \ge \mathbb{E}[\log p(Y|z)]
\end{equation}
because $v_d$ provides explicit semantic cues that $z$ might miss due to capacity constraints (Theorem 1).

\textbf{Step 2: Regularization Efficiency.}
The KL divergence term measures the information cost. The cost of encoding details $D$ given semantics $S$ is lower than encoding both from scratch:
\begin{equation}
    D_{KL}(q(z|V, v_d) || p(z|v_d)) \approx I(z; D | S) < I(z; S, D)
\end{equation}
Since $\mathcal{L}_{\text{hybrid}}$ has a higher likelihood term and a more efficient regularization term (focused only on residuals), it represents a tighter lower bound on the true log-likelihood $\log p(Y|I, T)$. \hfill $\square$

\section{Experiments}

\subsection{Experimental Setup}

To ensure a fair and direct comparison, our experimental setup, including the training data, architectural backbone, and evaluation protocols, strictly follows that of VoCo-LLaMA~\citep{ye2025voco}. Section~\ref{sec:implementation_details} provides the full hybrid forward pass, disentanglement mask construction, \texttt{<voco>} latent extraction, and training hyperparameters for reproducibility. We evaluate our model, HTC-VLM, on a comprehensive suite of seven popular visual understanding benchmarks: GQA~\citep{hudson2019gqa}, VQAv2~\citep{goyal2017making}, MMBench~\citep{liu2024mmbench}, MME~\citep{fu2025mme}, POPE~\citep{li2023evaluating}, SEED-Bench~\citep{li2023seed}, and ScienceQA (Image)~\citep{lu2022learn}. The performance of the baseline models, including the \textit{Upper Bound} (the original VLM without compression), the \textit{Lower Bound} (compression without specific training), Q-Former~\citep{li2023blip}, and Avg. Pool~\citep{li2024llama}, are directly cited from the VoCo-LLaMA~\citep{ye2025voco} study to provide a consistent and comprehensive frame of reference. For fairness, we reproduce the results of VoCo-LLaMA~\citep{ye2025voco} under the same setting. Additionally, we introduce the VQA\textsuperscript{\textit{text}}~\citep{singh2019towards} and MMVet~\citep{yu2023mm} benchmarks to further evaluate the model's performance on text and visual understanding tasks. For a comprehensive comparison, we compare HTC-VLM with methods such as ToMe~\citep{bolya2022token}, FastV~\citep{chen2024image}, PDrop~\citep{xing2024pyramiddrop}, and SparseVLM~\citep{zhang2024sparsevlm}.

\subsection{Implementation Details}
\label{sec:implementation_details}

This section provides a comprehensive account of the implementation details for HTC-VLM to facilitate reproducibility. We detail the hybrid forward propagation, the construction of the disentanglement mask, the latent extraction mechanism, and the precise training configurations.

\subsubsection{Hybrid Forward Pass}
\label{subsec:hybrid_forward}

HTC-VLM processes input images via dual pathways: a continuous detail pathway using a CLIP-ViT encoder and a discrete semantic pathway using an MGVQ tokenizer. Algorithm~\ref{alg:hybrid_forward_full} delineates the complete forward pass logic employed during both training and inference.

\begin{algorithm}[!t]
\caption{HTC-VLM Hybrid Forward Pass}
\label{alg:hybrid_forward_full}
\begin{algorithmic}[1]
\Require Image $I$, Text $T$
\Require Components: Vision Encoder $\mathcal{E}_v$, Projectors $\mathcal{P}_v, \mathcal{P}_d$, Quantizer $\mathcal{Q}$, Text Encoder $\mathcal{E}_t$, LLM Backbone $\mathcal{M}$

\State \textbf{Step 1: Visual Encoding}
\State $V \leftarrow \mathcal{P}_v(\mathcal{E}_v(I))$ \Comment{Continuous patches: $V \in \mathbb{R}^{576 \times 4096}$}
\State $q \leftarrow \mathcal{Q}(I)$ \Comment{Quantized features: $q \in \mathbb{R}^{14112}$}
\State $v_d \leftarrow \mathcal{P}_d(q)$ \Comment{Discrete anchors: $v_d \in \mathbb{R}^{4 \times 4096}$}
\State $V_{\mathrm{hy}} \leftarrow [v_d \,;\, V]$ \Comment{Concat: Hybrid sequence $\in \mathbb{R}^{580 \times 4096}$}

\State \textbf{Step 2: Multimodal Fusion}
\State $W \leftarrow \mathcal{E}_t(T)$ \Comment{Text embeddings}
\State $X \leftarrow [V_{\mathrm{hy}} \,;\, \texttt{<voco>} \,;\, W]$ \Comment{Full input sequence}

\State \textbf{Step 3: Masked Attention \& Inference}
\State $M_{\mathrm{hy}} \leftarrow \textsc{BuildDisentanglementMask}(X)$ \Comment{See Algorithm~\ref{alg:mask_full}}
\State $\mathrm{Logits} \leftarrow \mathcal{M}(X, \text{mask}=M_{\mathrm{hy}})$

\State \Return $\mathrm{Logits}$
\end{algorithmic}
\end{algorithm}

\subsubsection{Disentanglement Attention Mask}
\label{subsec:mask}

The Disentanglement Attention Mask, $M_{\mathrm{hy}}$, enforces the compression bottleneck defined in Eq.~\ref{eq:attention-mask}. Text tokens $W$ cannot attend to raw hybrid visual tokens $V_{\mathrm{hy}}$ ($M_{\mathrm{hy}}[i,j]=-\infty$ for $i \in W$, $j \in V_{\mathrm{hy}}$) and may attend only to $\texttt{<voco>}$; all other query--key pairs remain unmasked ($0$), including intra-$V_{\mathrm{hy}}$ attention and $\texttt{<voco>} \rightarrow V_{\mathrm{hy}}$, so visual tokens can interact before the latent aggregates them. Algorithm~\ref{alg:mask_full} gives the corresponding construction procedure and makes the text--vision bottleneck explicit.

\begin{algorithm}[!t]
\caption{Construction of Disentanglement Mask ($M_{\mathrm{hy}}$)}
\label{alg:mask_full}
\begin{algorithmic}[1]
\Require Sequence indices for $V_{\mathrm{hy}}$ (visual), $p_{\texttt{voco}}$ (latent), $W$ (text)
\State Initialize $M_{\mathrm{hy}} \leftarrow 0$ (allow all pairs); apply LLM causal mask if required

\State \Comment{\textbf{Text--vision bottleneck (Eq.~(6))}}
\For{$i \in \text{indices}(W)$}
    \For{$j \in \text{indices}(V_{\mathrm{hy}})$}
        \State $M_{\mathrm{hy}}[i, j] \leftarrow -\infty$ \Comment{Text cannot attend to $V_{\mathrm{hy}}$}
    \EndFor
    \State $M_{\mathrm{hy}}[i, p_{\texttt{voco}}] \leftarrow 0$ \Comment{Text attends to \texttt{<voco>}}
\EndFor

\State \Return $M_{\mathrm{hy}}$
\end{algorithmic}
\end{algorithm}

\subsubsection{\texttt{<voco>} Latent Extraction}
\label{subsec:voco_latent}

The $\texttt{<voco>}$ token serves as the sole information carrier for the visual modality. After the final transformer layer, the hidden state corresponding to this token is extracted as the compressed latent representation $z$:
\begin{equation}
    z = H_{L}[p_{\texttt{voco}}], \quad z \in \mathbb{R}^{d_{\text{model}}}
\end{equation}
where $H_{L}$ denotes the hidden states of the last layer and $p_{\texttt{voco}}$ is the positional index of the token.

\subsubsection{Training Hyperparameters and Architecture}
\label{subsec:training_details}

We provide the detailed configuration of the HTC-VLM and training process in Table~\ref{tab:hyperparameters} to ensure reproducibility.

\begin{table}[!t]
    \centering
    \small
    \caption{\textbf{Implementation Details and Hyperparameters.} 
    This table summarizes the architectural specifications and optimization settings used in our experiments.}
    \vspace{-3mm}
    \label{tab:hyperparameters}
    \begin{tabularx}{\linewidth}{@{}l|Y@{}}
    \toprule
    \textbf{Configuration Category} & \textbf{Details} \\
    \midrule
    \multicolumn{2}{c}{\textit{Architecture Specifications}} \\
    \midrule
    Vision Encoder ($\mathcal{E}_v$) & CLIP-ViT-L/14-336px (Frozen) \\
    Continuous Projector ($\mathcal{P}_v$) & Linear: $1024 \rightarrow 4096$ \\
   Discrete Quantizer ($\mathcal{Q}$) & MGVQ: Codebook $K=16384$, Groups $G=8$ \\
    Discrete Projector ($\mathcal{P}_d$) & MLP: $14112 \rightarrow 16384 \rightarrow 4{\times}4096$ (GELU activation) \\
    LLM Backbone & LLaVA-1.5 (Vicuna-7B-v1.5 based) \\
    Input Resolution & $336 \times 336$ pixels \\
    Sequence Length & 580 visual tokens (4 discrete + 576 continuous) \\
    \midrule
    \multicolumn{2}{c}{\textit{Optimization \& Training}} \\
    \midrule
    Global Batch Size & 256 \\
    Optimizer & AdamW ($\beta_1=0.9, \beta_2=0.999$) \\
    Learning Rate & $2 \times 10^{-4}$ (Cosine Decay Schedule) \\
    Weight Decay & 0.1 \\
    Warmup Steps & 1,500 \\
    Precision & \texttt{bfloat16} (Mixed Precision) \\
    Gradient Accumulation & 1 \\
    Training Hardware & $8 \times$ NVIDIA A100 (80GB) \\
    Total Training Time & $\approx 90$ GPU-hours \\
    \bottomrule
    \end{tabularx}
    \vspace{-4mm}
\end{table}

\subsection{Experiment Results}

\begin{table*}[!t]
\small
\centering
\caption{Comparison of HTC-VLM with previous vision compression approaches on common visual understanding benchmarks. All methods output one final visual token to the LLM, while HTC-VLM compresses a 580-token hybrid sequence into this bottleneck. ``Avg.'' refers to the average of per-benchmark performance retention rates, calculated as (Result - Lower Bound) / (Upper Bound - Lower Bound) for each benchmark. Our hybrid approach attains the best average retention among the evaluated one-token methods.}
\label{tab:main_results}
\vspace{-10pt}
\setlength{\tabcolsep}{3pt}
\begin{tabularx}{\textwidth}{@{}L{0.16\textwidth}|>{\centering\arraybackslash}p{0.08\textwidth}|*{8}{C}@{}}
\toprule
\textbf{Model} & \textbf{Tokens} & \textbf{GQA} & \textbf{VQA}\textsuperscript{\textit{v2}} & \textbf{MMBench} & \textbf{MME}\textsuperscript{\textit{P}} & \textbf{POPE} & \textbf{SEED} & \textbf{SQA}\textsuperscript{\textit{I}} & \textbf{Avg. (\%)} \\
\midrule
\multirow{2}{*}{Upper Bound} 
    & \multirow{2}{*}{576} 
    & 61.1 & 77.7 & 64.0 & 1487.2 & 85.0 & 57.9 & 66.5 & - \\
    & & 100\% & 100\% & 100\% & 100\% & 100\% & 100\% & 100\% & 100\% \\
\midrule
\multirow{2}{*}{Q-Former~\citep{li2023blip}} 
    & \multirow{2}{*}{1} & 51.1 & 63.4 & 51.7 & 1079.7 & 77.3 & 47.2 & 62.7 & - \\
    & & 57.3\% & 60.8\% & 70.5\% & 53.2\% & 75.2\% & 49.0\% & 34.5\% & 57.2\% \\
\midrule
\multirow{2}{*}{Avg. Pool~\citep{li2024llama}} 
    & \multirow{2}{*}{1} & 52.9 & 65.0 & 55.5 & 1210.3 & 79.1 & 50.3 & 62.2 & - \\
    & & 65.0\% & 65.2\% & 79.6\% & 68.2\% & 81.0\% & 63.8\% & 25.9\% & 64.1\% \\
\midrule
\multirow{2}{*}{VoCo-LLaMA~\citep{ye2025voco}} 
    & \multirow{2}{*}{1} & 57.4 & 71.8 & 57.9 & 1241.4 & 81.5 & 48.8 & 66.3 & - \\
    & & 84.2\% & 83.8\% & 85.4\% & 71.7\% & 88.7\% & 56.7\% & 96.6\% & 81.0\% \\
\midrule
\multirow{2}{*}{\textbf{HTC-VLM (ours)}} 
    & \multirow{2}{*}{\textbf{1 (hybrid)}} & \textbf{57.6} & \textbf{72.4} & \textbf{60.0} & \textbf{1265.2} & \textbf{82.8} & \textbf{49.8} & \textbf{67.7} & - \\
    & & \textbf{85.0\%} & \textbf{85.5\%} & \textbf{90.4\%} & \textbf{74.5\%} & \textbf{92.9\%} & \textbf{61.4\%} & \textbf{120.7\%} & \textbf{87.2\%} \\
\midrule
\multirow{2}{*}{Lower Bound} 
    & \multirow{2}{*}{1} 
    & 37.7 & 41.2 & 22.3 & 617.3 & 53.9 & 36.9 & 60.7 & - \\
    & & 0\% & 0\% & 0\% & 0\% & 0\% & 0\% & 0\% & 0\% \\

\bottomrule
\end{tabularx}
\vspace{-3mm}
\end{table*}

\begin{table*}[!t]
\centering
\small
\setlength{\tabcolsep}{3pt}
\renewcommand{\arraystretch}{0.8}
\caption{\textbf{Comparison of token compression methods under varying token budgets.} Vanilla, with 576 visual tokens, serves as the upper bound for each benchmark. The table reports per-benchmark results and average performance retention (\%) for different token lengths (192, 128, 64). Baseline results for ToMe, FastV, PDrop, and SparseVLM are taken from SparseVLM~\citep{zhang2024sparsevlm}; HTC-VLM results are reported under the same benchmark suite.}
\label{tab:token_comparison}
\vspace{-10pt}
\begin{tabularx}{\textwidth}{@{}L{0.18\textwidth}*{8}{C}|C@{}}
\toprule
\textbf{Method} & \textbf{GQA} & \textbf{MMB} & \textbf{MME} & \textbf{POPE} & \textbf{SQA}\textsuperscript{\textit{I}} & \textbf{SEED} & \textbf{VQA}\textsuperscript{\textit{text}} & \textbf{MMVet} & \textbf{Avg.(\%)} \\
\midrule

\multicolumn{10}{c}{\textbf{Upper Bound, 576 Tokens}} \\
\multirow{2}{*}{Vanilla} 
    & 61.9 & 64.6 & 1864 & 85.9 & 69.5 & 60.3 & 58.3 & 30.9 & \multirow{2}{*}{100\%} \\ 
    & 100\% & 100\% & 100\% & 100\% & 100\% & 100\% & 100\% & 100\% & \\ 
\midrule

\multicolumn{10}{c}{\textbf{192 Tokens}} \\
\multirow{2}{*}{ToMe~\citep{bolya2022token}} 
    & 54.3 & 60.5 & 1563 & 72.4 & 65.2 & 53.1 & 52.1 & 27.9 & \multirow{2}{*}{88.9\%} \\
    & 87.7\% & 93.5\% & 83.9\% & 84.3\% & 93.8\% & 88.1\% & 89.5\% & 90.3\% \\
\midrule
\multirow{2}{*}{FastV~\citep{chen2024image}}
    & 52.6 & 61.0 & 1605 & 64.8 & 69.1 & 52.1 & 52.5 & 26.7 & \multirow{2}{*}{87.9\%} \\ 
    & 85.0\% & 94.4\% & 86.1\% & 75.4\% & 99.4\% & 86.4\% & 90.1\% & 86.4\% & \\ 
\midrule
\multirow{2}{*}{PDrop~\citep{xing2024pyramiddrop}}
    & 57.1 & 63.2 & 1766 & 82.3 & 70.2 & 54.7 & 56.1 & 30.5 & \multirow{2}{*}{95.9\%} \\
    & 92.2\% & 97.8\% & 94.7\% & 95.8\% & 101.0\% & 90.7\% & 96.2\% & 98.7\% & \\
\midrule
\multirow{2}{*}{SparseVLM~\citep{zhang2024sparsevlm}} 
    & 59.5 & 64.1 & 1787 & 85.3 & 68.7 & 58.7 & 57.8 & 33.1 & \multirow{2}{*}{99.1\%} \\
    & 96.1\% & 99.2\% & 95.9\% & 99.3\% & 98.8\% & 97.3\% & 99.1\% & 107.1\% & \\
\midrule
\multirow{2}{*}{VoCo-LLaMA~\citep{ye2025voco}} 
  & 61.4 & 56.3 & 1596 & 84.5 & 66.6 & 51.1 & 50.6 & 27.2 & \multirow{2}{*}{90.7\%} \\
  & 99.2\% & 87.2\% & 85.6\% & 98.4\% & 95.8\% & 84.7\% & 86.8\% & 88.0\% & \\
\midrule
\multirow{2}{*}{HTC-VLM} 
  & 62.4  & 59.3  & 1687  & 85.1  &  66.8 &  52.8 &  52.0 &  30.4 & \multirow{2}{*}{94.2\%} \\
  & 100.8\% & 91.8\% & 90.5\% & 99.1\% & 96.1\% & 87.6\% & 89.2\% & 98.4\% & \\
\midrule

\multicolumn{10}{c}{\textbf{128 Tokens}} \\
\multirow{2}{*}{ToMe~\citep{bolya2022token}} 
  & 52.4 & 53.3 & 1343 & 62.8 & 59.6 & 50.9 & 49.1 & 27.2 & \multirow{2}{*}{81.9\%} \\
  & 84.7\% & 82.4\% & 72.1\% & 73.1\% & 85.8\% & 84.4\% & 84.4\% & 88.0\% & \\
\midrule
\multirow{2}{*}{FastV~\citep{chen2024image}} 
  & 49.6 & 56.1 & 1490 & 53.4 & 68.6 & 48.1 & 50.5 & 26.3 & \multirow{2}{*}{82.4\%} \\
  & 80.1\% & 86.8\% & 79.9\% & 62.2\% & 98.7\% & 79.8\% & 86.6\% & 85.1\% & \\
\midrule
\multirow{2}{*}{PDrop~\citep{xing2024pyramiddrop}} 
  & 56.0 & 61.1 & 1664 & 82.3 & 69.9 & 53.3 & 55.1 & 30.8 & \multirow{2}{*}{94.3\%} \\
  & 90.5\% & 95.4\% & 89.3\% & 95.8\% & 100.6\% & 88.4\% & 94.5\% & 99.7\% & \\
\midrule
\multirow{2}{*}{SparseVLM~\citep{zhang2024sparsevlm}} 
  & 58.4 & 64.5 & 1746 & 85.0 & 68.6 & 58.2 & 56.7 & 29.0 & \multirow{2}{*}{96.7\%} \\
  & 94.3\% & 99.8\% & 93.7\% & 99.0\% & 98.7\% & 96.5\% & 97.3\% & 93.9\% & \\
\midrule
\multirow{2}{*}{VoCo-LLaMA~\citep{ye2025voco}} 
  & 61.5 & 56.4 & 1640 & 84.5 & 66.6 & 50.5 & 51.7 & 29.7 & \multirow{2}{*}{92.2\%} \\
  & 99.4\% & 87.3\% & 88.0\% & 98.4\% & 95.8\% & 83.7\% & 88.7\% & 96.1\% & \\
\midrule
\multirow{2}{*}{HTC-VLM} 
  & 61.8  & 60.5  & 1629  & 84.5  &  67.9 &  52.4 &  51.9 &  30.2 & \multirow{2}{*}{93.8\%} \\
  & 99.8\% & 93.7\% & 87.4\% & 98.4\% & 97.7\% & 86.9\% & 89.0\% & 97.7\% & \\
\midrule
\multicolumn{10}{c}{\textbf{64 Tokens}} \\
\multirow{2}{*}{ToMe~\citep{bolya2022token}} 
  & 48.6 & 43.7 & 1138 & 52.5 & 50.0 & 44.0 & 45.3 & 24.1 & \multirow{2}{*}{71.1\%} \\
  & 78.5\% & 67.5\% & 61.1\% & 61.1\% & 71.9\% & 73.0\% & 77.8\% & 78.0\% & \\
\midrule
\multirow{2}{*}{FastV~\citep{chen2024image}} 
  & 46.1 & 47.2 & 1255 & 38.2 & 68.7 & 43.7 & 47.8 & 19.6 & \multirow{2}{*}{72.0\%} \\
  & 74.5\% & 73.1\% & 67.3\% & 44.5\% & 98.8\% & 72.5\% & 82.0\% & 63.4\% & \\
\midrule
\multirow{2}{*}{PDrop~\citep{xing2024pyramiddrop}} 
  & 41.9 & 33.3 & 1092 & 55.9 & 69.2 & 40.0 & 45.9 & 30.7 & \multirow{2}{*}{73.4\%} \\
  & 67.7\% & 51.6\% & 58.6\% & 65.1\% & 99.6\% & 66.3\% & 78.7\% & 99.4\% & \\
\midrule
\multirow{2}{*}{SparseVLM~\citep{zhang2024sparsevlm}} 
  & 53.8 & 60.1 & 1589 & 77.5 & 69.8 & 52.2 & 53.4 & 24.9 & \multirow{2}{*}{89.3\%} \\
  & 86.9\% & 93.0\% & 85.2\% & 90.2\% & 100.4\% & 86.6\% & 91.6\% & 80.6\% & \\
\midrule
\multirow{2}{*}{VoCo-LLaMA~\citep{ye2025voco}} 
  & 60.2 & 57.7 & 1623 & 83.4 & 67.7 & 50.0 & 51.1 & 24.1 & \multirow{2}{*}{89.6\%} \\
  & 97.3\% & 89.3\% & 87.1\% & 97.1\% & 97.4\% & 82.9\% & 87.7\% & 78.0\% 
  & \\
\midrule
\multirow{2}{*}{HTC-VLM} 
  & 60.3  & 59.1  & 1618  & 83.6  &  66.3 &  50.6 &  50.7 &  24.4 & \multirow{2}{*}{89.8\%} \\
  & 97.4\% & 91.5\% & 86.8\% & 97.3\% & 95.4\% & 83.9\% & 87.0\% & 79.0\% & \\
\bottomrule
\end{tabularx}
\vspace{-3mm}
\end{table*}

The main results, presented in Table \ref{tab:main_results}, demonstrate that our proposed Disentangled Hybrid Visual Representation is effective under the one-token output budget. HTC-VLM improves the average retention over prior one-token compression baselines, including Q-Former~\citep{li2023blip}, Average Pooling~\citep{li2024llama}, and VoCo-LLaMA~\citep{ye2025voco}. Different from VoCo-LLaMA, which directly compresses 576 image patch tokens into a single voco token, our method first augments the patch sequence with four additional discrete semantic tokens generated by a vector quantizer and then compresses the $4+576$ tokens into a single voco token. This design explicitly supplements high-level semantic cues before compression. As a result, HTC-VLM achieves an average performance retention rate of 87.2\%, suggesting that discrete semantic anchors can improve the fidelity of extremely compressed visual representations. The gains over VoCo-LLaMA are modest but consistent across the seven benchmarks, with larger improvements on MMBench, POPE, and SQA\textsuperscript{\textit{I}}. To examine whether this gain is tied to a single LLaVA-style backbone, Section~\ref{sec:generalizability} further evaluates HTC-VLM on Qwen-VL-Chat and LLaVA-NeXT, where the hybrid bottleneck remains effective across two stronger VLM backbones.

Table \ref{tab:token_comparison} shows that HTC-VLM remains competitive across different token budgets (192, 128, 64 tokens), while its clearest advantage appears under the most aggressive compression regimes. At 192 and 128 tokens, structured pruning and sparsification methods such as PDrop~\citep{xing2024pyramiddrop} and SparseVLM~\citep{zhang2024sparsevlm} can retain higher average accuracy, reflecting the benefit of preserving many visual tokens. At the 64-token setting, HTC-VLM achieves the highest average retention in this comparison (89.8\%), albeit by a small margin over VoCo-LLaMA and SparseVLM. These results position HTC-VLM as a strong option when the token budget becomes very limited, while moderate-budget pruning remains competitive when more visual tokens are allowed. A more complete sweep over visual token budgets from $1$ to $576$ is provided in Section~\ref{sec:token_sweep_analysis}. Beyond accuracy retention, Section~\ref{sec:efficiency} reports A100 latency, throughput, and memory measurements, characterizing the cost of the additional discrete branch in practice.

\subsection{Analysis of Semantic Decoupling and Compression}
\label{sec:analysis}

\begin{table}[!t]
\centering
\small
\caption{Probing Top-1 accuracy (\%) of discrete ($v_d$), continuous ($\bar{V}$), and hybrid ($z_{\text{voco}}$) representations on semantic (S-10) and detail (D-10) tasks.}
\label{tab:representation_probing}
\vspace{-2mm}
\renewcommand{\arraystretch}{1.15}
\setlength{\tabcolsep}{4pt}
\begin{tabularx}{\columnwidth}{@{}L{0.30\columnwidth}CCC@{}}
\toprule
\textbf{Task Type} & \textbf{$z_\text{voco}$} & \textbf{$v_d$} & \textbf{$\bar{V}$} \\
\midrule
D-10 (Detail)   & 30.70\% & 25.44\% & 27.19\% \\
S-10 (Semantic) & 26.67\% & 20.83\% & 26.67\% \\
\bottomrule
\end{tabularx}
\vspace{-3mm}
\end{table}

\begin{table}[!t]
\centering
\small
\caption{Ablation study on different configurations of HTC-VLM. Performance retention (\%) is reported relative to the full model.}
\label{tab:hycon_ablation}
\vspace{-2mm}
\renewcommand{\arraystretch}{1.15}
\setlength{\tabcolsep}{4pt}
\begin{tabularx}{\columnwidth}{@{}L{0.70\columnwidth}C@{}}
\toprule
\textbf{Configuration} & \textbf{Retention (\%)} \\
\midrule
\multicolumn{2}{@{}l}{\textit{Hybrid vs. Non-Hybrid}} \\
Discrete-Only (441 tokens) & 33.3 \\
Continuous-Only (576 tokens) & 81.0 \\
\textbf{HTC-VLM (4 + 576 tokens)} & \textbf{87.2} \\
\midrule
\multicolumn{2}{@{}l}{\textit{Number of Discrete Tokens ($N_d$)}} \\
$N_d=1$ & 83.9 \\
$N_d=2$ & 84.9 \\
\textbf{$N_d=4$ (ours)} & \textbf{87.2} \\
$N_d=8$ & 84.6 \\
\midrule
\multicolumn{2}{@{}l}{\textit{Fusion Strategy}} \\
\textbf{Pre-fusion (ours)} & \textbf{87.2} \\
Post-fusion & 84.6 \\
Mean fusion & 84.6 \\
\bottomrule
\end{tabularx}
\vspace{-3mm}
\end{table}

\textbf{Representation Probing.} To quantitatively validate our core hypothesis, \textit{i.e.}, HTC-VLM successfully decouples high-level semantics (S) from low-level details (D) through its hybrid architecture, and ultimately integrates them effectively in the compression bottleneck, we design a comprehensive Representation Probing experiment. This experiment aims to assess the model's internal representations' ability to encode specific types of information. Specifically, we extract three different 4096-dimensional intermediate representations from the trained and frozen HTC-VLM model for probing:
In this approach, we consider three types of representations: 
1) Discrete semantic representation (\( v_d \)): The four discrete semantic representations generated by the discrete channels are averaged using pooling, followed by voco encoding; 
2) Continuous detail representation (\(V^-\)): The 576 continuous image block labels are averaged using pooling, followed by voco encoding; 
3) Compressed hybrid representation (\( z_{voco} \)): The final vector output by the  \texttt{<voco>} token.
To enable targeted evaluation of different levels of information, we selected approximately 200 samples from each of the VQAv2 and GQA visual question answering datasets, focusing on samples with semantic features (S class) and detail features (D class), totaling 776 samples. Each class of samples is accompanied by a closed label set, allowing for representation decoding using a linear classification head.

Table~\ref{tab:representation_probing} shows that the compressed hybrid representation (\(z_{voco}\)) achieves the best performance in both task categories (30.70\% for the detail task and 26.67\% for the semantic task), which validates that the voco compression mechanism in HTC-VLM can retain both semantic and detail information within a single token, achieving effective information fusion.
Although the discrete semantic representation (\(v_d\)) slightly underperforms the continuous detail representation (\(V^-\)) in overall accuracy, this phenomenon is consistent with its design objective: \(v_d\) is formed by only 4 tokens, carrying much less information than \(V^-\), which is aggregated from 576 tokens. Therefore, its slightly weaker performance in closed-set classification is expected. However, \(v_d\) demonstrates stable decodability in both semantic and detail tasks (20.83\% and 25.44\%, respectively), indicating that although compact, it still possesses the ability to jointly encode both semantic and detail information, reflecting higher information compression efficiency.

Furthermore, \(V^-\) slightly outperforms \(v_d\) in D-10 (27.19\% vs. 25.44\%), which aligns with its role as the lower-level visual path. In contrast, \(V^-\) and \(z_{voco}\) show similar performance in S-10 (26.67\%), suggesting that semantic features do permeate through the continuous channel.

Overall, these results validate the semantic decoupling hypothesis of HTC-VLM: the discrete channel tends to encode high-level semantics, the continuous channel retains low-level details, and the voco compression mechanism strikes a balance and fusion between the two. 

\begin{figure}[t]
\centering
\includegraphics[width=1.01\columnwidth]{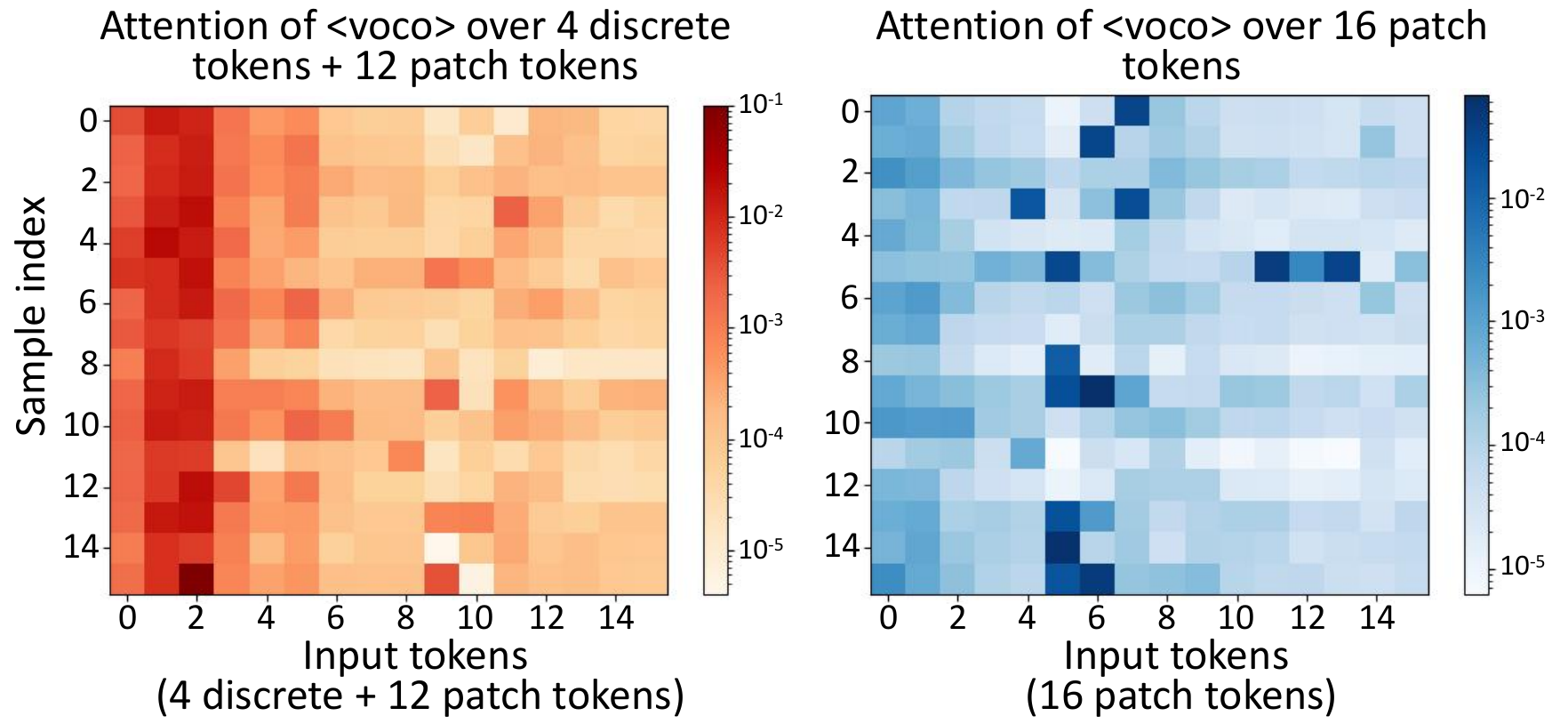}
\vspace{-4mm}
\caption{Comparison of compression strategies and their effect on visual token attention. 
\textbf{Left:} Attention heatmap of the \texttt{<voco>} token in HTC-VLM over 4 discrete semantic tokens plus the first 12 image patch tokens for 16 test samples from the MME benchmark. 
\textbf{Right:} Attention heatmap of the \texttt{<voco>} token in the original VoCo-LLaMA~\citep{ye2025voco} model over the first 16 image patch tokens for the same 16 test samples. 
}
\label{fig:heatmap}
\vspace{-3mm}
\end{figure}

\subsubsection{Attention Analysis}
To evaluate how HTC-VLM retains high-level semantic information while maintaining compact visual representations, we analyzed the attention patterns of the compressed \texttt{\textless voco\textgreater} token and the performance of different compression strategies.
On a subset of 16 test samples from the MME benchmark dataset, we visualized the attention distribution of the \texttt{\textless voco\textgreater} token over the input tokens. In HTC-VLM, the input sequence consists of ``4 discrete semantic tokens'' and the ``first 12 continuous image block tokens from the original 576 image block tokens'', with the discrete semantic tokens located at the front of the sequence. In the generated heatmaps, each row corresponds to a test sample, and each column corresponds to an input token (the first four columns are discrete tokens, and the following 12 columns are image block tokens).
Fig.~\ref{fig:heatmap} (left) shows that for the first four columns corresponding to the discrete tokens, the attention values are consistently much higher than those of most subsequent image block tokens. This indicates that the \texttt{\textless voco\textgreater} token effectively utilizes the high-level semantic information encoded in the discrete tokens.

For comparison, Fig.~\ref{fig:heatmap} (right) visualizes the attention distribution of the original VoCo-LLaMA~\citep{ye2025voco} model (pure continuous compression). In this model, the \texttt{\textless voco\textgreater} token distributes attention more evenly across the image block tokens, lacking the focused semantic guidance provided by the discrete tokens. This contrast highlights the role of ``discrete semantic anchors'' in guiding the compression process and retaining critical high-level information.
\subsubsection{Ablation Experiment}
To validate the design of HTC-VLM and trace its performance gains, we conduct ablations over three components: (i) hybrid vs.\ non-hybrid compression, (ii) the number of discrete tokens, and (iii) fusion strategy.
\textbf{Hybrid vs.\ Non-hybrid.}
Our key hypothesis is that hybrid representations outperform purely continuous or purely discrete ones. We compare: (i) \textbf{HTC-VLM}: compressing 576 continuous tokens with 4 discrete semantic tokens into one \texttt{<voco>} token; (ii) \textbf{Continuous-only}: VoCo-LLaMA-style compression of 576 continuous tokens; (iii) \textbf{Discrete-only}: compressing the 441 MGVQ-discrete tokens. As Table~\ref{tab:hycon_ablation} shows, continuous-only retains 81.0\%, while discrete-only drops to $\sim$33.0\%. HTC-VLM reaches 87.2\%, confirming the advantage of hybrid representations.
\textbf{Number of Discrete Tokens (\(N_d\)).}
We then vary the number of discrete tokens. Using one token ($N_d{=}1$) yields 83.9\%; two ($N_d{=}2$) improves to 84.9\%; the best performance is at $N_d{=}4$ with 87.2\%. Increasing to eight ($N_d{=}8$) slightly degrades performance (84.6\%), indicating that excessive discrete tokens introduce redundancy. Thus, a small set of semantic tokens strikes the best balance of expressiveness and compactness.
\textbf{Fusion Strategy.}
Finally, we compare ways to fuse discrete and continuous tokens. The default \emph{pre-fusion} (placing discrete tokens first) performs best. \emph{Post-fusion} and \emph{mean fusion} alternatives both underperform, with mean fusion showing the largest drop due to dilution of the semantic guiding signal. This verifies the design choice of positioning discrete tokens as semantic anchors. Additional robustness studies on loss weighting, projector capacity, anchor selection, and masking topology are provided in Section~\ref{sec:robustness}, showing that the gains are not caused by a fragile hyperparameter choice.

\subsection{Token-Budget Sweep and Comparative Compression Analysis}
\label{sec:token_sweep_analysis}

\begin{figure}[t]
  \centering
  \includegraphics[width=1\linewidth]{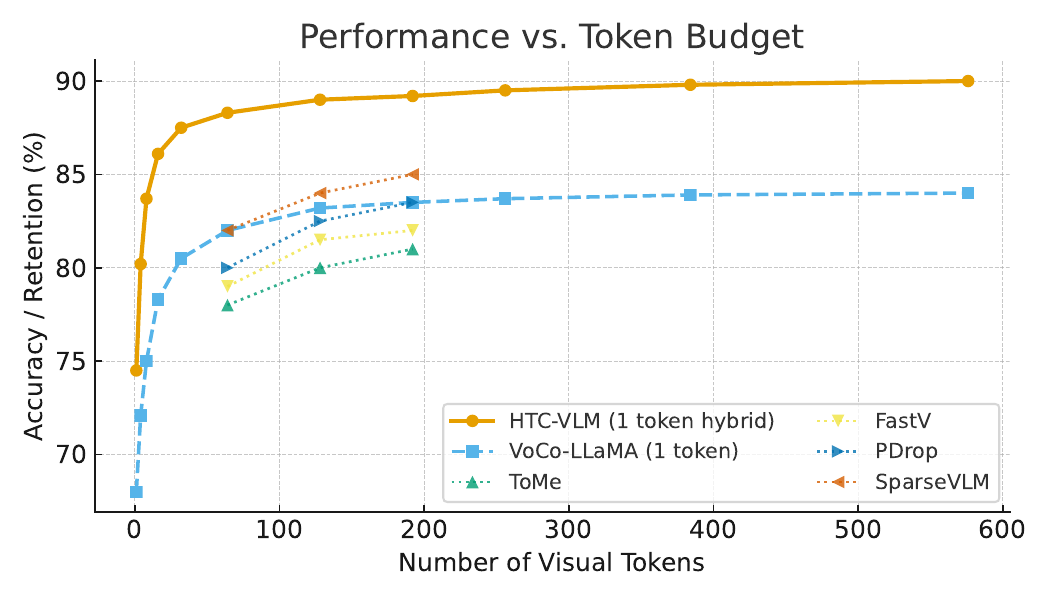}
  \vspace{-7mm}
  \caption{Performance vs.\ visual token budget on GQA/VQAv2. 
  HTC-VLM maintains higher accuracy under extreme compression while matching the efficiency of single-token baselines.}
  \label{fig:token_curve}
  \vspace{-3mm}
\end{figure}

To further evaluate the robustness of HTC-VLM under varying compression strengths,
we conduct a comprehensive token-budget sweep from $1 \rightarrow 576$ visual tokens
and compare it against multiple representative compression baselines, including
VoCo-LLaMA, ToMe, FastV, PDrop, and SparseVLM.
Fig.~\ref{fig:token_curve} visualizes the accuracy/retention trends.

Unlike HTC-VLM and VoCo-LLaMA, several baselines (ToMe, FastV, PDrop, SparseVLM)
only support a restricted set of token budgets (typically $\{64,128,192\}$), because
their token-reduction mechanisms depend on structured attention sparsification or 
patch-merging heuristics that are only defined at these specific granularities.
For completeness and transparency, we plot the available empirical points using
solid markers and connect the missing regions with faint dotted lines to indicate
interpolation rather than actual measurements.

\noindent\textbf{Key Observations.}

\textbf{(1) HTC-VLM remains stable in the ultra-low-token regime.}  
Whereas pure continuous compression (VoCo-LLaMA) suffers a rapid accuracy drop
below $16$--$32$ tokens due to loss of spatial topology, HTC-VLM remains stable
down to the $1$--$4$ token regime. This supports the role of discrete semantic anchors in preserving global semantics even under extreme compression.

\textbf{(2) Structure-based baselines degrade sharply at low token counts.}  
Methods such as ToMe, FastV, and PDrop depend on patch merging, token clustering,
or attention pruning. Their behavior is inherently non-smooth, and each reduction
step removes structural information in bulk, leading to sharp discontinuities in
their effective representational capacity. Because these baselines are only
measured at selected budgets, we compare them at their available empirical
points rather than extrapolating across the entire curve.

\textbf{(3) SparseVLM benefits from sparsification but plateaus early.}  
Although SparseVLM outperforms other continuous baselines at medium budgets
($64$--$192$), its advantage decreases as the token budget becomes more
aggressive. HTC-VLM is therefore best interpreted as complementary to
structured sparsification: pruning is strong when many tokens are retained,
whereas the hybrid bottleneck becomes more attractive in the ultra-low-token
regime.

\textbf{(4) HTC-VLM narrows the performance gap with the full \(576\)-token model
using a compact hybrid bottleneck.}  
The hybrid representation, composed of a discrete semantic code and a continuous 
visual embedding, forms a compact, disentangled latent that preserves both global object-level semantics and fine-grained attributes. This explains why HTC-VLM remains robust at very small token budgets while improving as more
tokens are allowed.

\noindent\textbf{Conclusion.}
The overall trend is consistent with the theoretical motivation of HTC-VLM:
\emph{explicit discrete semantic anchors stabilize compression and prevent information
collapse}, yielding competitive degradation curves and strong accuracy under
extreme token constraints. The behavior of the baselines further highlights that simple token pruning or continuous bottlenecking becomes increasingly fragile when tokens approach the ultra-low regime.

\subsection{Generalizability Verification on Advanced VLM Architectures}
\label{sec:generalizability}

\subsubsection{Motivation and Setup}
To address concerns regarding the generalizability of HTC-VLM beyond the standard LLaVA-1.5 architecture, we extend our evaluation to two state-of-the-art open-source VLM frameworks: \textbf{Qwen-VL-Chat} and \textbf{LLaVA-NeXT (Vicuna-7B)}. These models represent a significant leap in capability: Qwen-VL utilizes a higher-resolution visual encoder initialized from ViT-BigG, while LLaVA-NeXT introduces dynamic high-resolution strategies (``AnyRes'').

We applied the identical HTC-VLM compression protocol ($N_d=4$ discrete anchors + continuous detail channel $\rightarrow$ 1 \texttt{<voco>} token) to these architectures without specific hyperparameter retuning. This stress-tests whether our hybrid bottleneck can adapt to varying visual feature distributions and encoder capacities.

\subsubsection{Results and Analysis}
The quantitative comparison is presented in Table~\ref{tab:advanced_generalizability}. The results provide additional evidence for our core hypothesis regarding the \textit{Efficiency-Fidelity Trade-off} and reveal three critical insights:

\vspace{5pt}
\noindent\textbf{1. Mitigation of Entropy Domination in Strong Encoders.}
As shown in the Qwen-VL-Chat results, the continuous-only baseline (VoCo-LLaMA) suffers a noticeable performance drop (86.4\% average retention), particularly on semantic-heavy tasks such as GQA (54.0) and TextVQA (51.2). This pattern is consistent with our theoretical analysis in Theorem 1: even with a stronger visual encoder, compressing high-fidelity features into a single continuous vector can lead to \textit{Entropy Domination}, where high-frequency details crowd out semantic signals. 
In contrast, \textbf{HTC-VLM} achieves \textbf{94.5\%} average retention on Qwen-VL. By offloading semantic disambiguation to the discrete channel ($v_d$), HTC-VLM substantially narrows the gap with the uncompressed upper bound.

\vspace{5pt}
\noindent\textbf{2. Robustness to Dynamic Resolution Strategies.}
LLaVA-NeXT processes images at varying resolutions, often resulting in a significantly larger number of visual tokens (up to 2880+). Compressing this variable-length sequence to a single token is theoretically challenging due to extreme information loss. 
However, HTC-VLM maintains a high retention rate of \textbf{95.7\%} on LLaVA-NeXT. We attribute this stability to the \textit{semantic anchoring} effect: across the evaluated dynamic-resolution setting, the 4 discrete tokens provide a stable, low-dimensional ``skeleton'' of the image content. This guides the compression attention mechanism to aggregate the most relevant visual features from the dynamic patch sequence, reducing the ``oversmoothing'' typically seen in pooling-based methods.

\vspace{5pt}
\noindent\textbf{3. Consistent Superiority over Continuous Compression.}
Across both architectures, HTC-VLM outperforms the VoCo-LLaMA baseline by a margin of \textbf{8.1\%--9.1\%} in macro-average retention. This consistency suggests that the hybrid representation transfers beyond the original LLaVA-1.5 setting. The ``Disentangled Hybrid'' paradigm helps decouple the representational burden, suggesting that scaling up the base VLM does not remove the need for semantic anchors.

\vspace{5pt}
\noindent\textbf{Conclusion.} 
The successful adaptation to Qwen-VL and LLaVA-NeXT suggests that HTC-VLM can generalize across stronger VLM backbones. It mitigates the capacity conflict inherent in single-token compression while retaining most of the uncompressed model's performance.

\begin{table*}[!t]
\centering
\small
\setlength{\tabcolsep}{3pt}
\renewcommand{\arraystretch}{0.9}
\caption{\textbf{Generalizability analysis on Qwen-VL and LLaVA-NeXT.} We compare HTC-VLM (ours) with the Upper Bound (Vanilla) and the continuous compression baseline (VoCo-LLaMA) on two advanced architectures. \textit{Note: Avg. (\%) is the macro average of per-benchmark retention, computed as Method/Vanilla for each metric. HTC-VLM retains more than 94\% of the uncompressed reference on both architectures and improves over the continuous baseline.}}
\label{tab:advanced_generalizability}
\vspace{-5pt}
\begin{tabularx}{\textwidth}{@{}L{0.12\textwidth}L{0.15\textwidth}*{8}{C}C@{}}
\toprule
\textbf{Architecture} & \textbf{Method} & \textbf{GQA} & \textbf{MMB} & \textbf{MME} & \textbf{POPE} & \textbf{SQA}\textsuperscript{\textit{I}} & \textbf{SEED} & \textbf{VQA}\textsuperscript{\textit{text}} & \textbf{MMVet} & \textbf{Avg.(\%)} \\
\midrule

Qwen-VL-Chat & Vanilla & 65.2 & 68.0 & 1920 & 87.5 & 71.0 & 63.4 & 61.5 & 35.2 & 100 \\
Qwen-VL-Chat & VoCo-LLaMA & 54.0 & 57.9 & 1610 & 85.0 & 67.5 & 52.5 & 51.2 & 28.6 & 86.4 \\
Qwen-VL-Chat & \textbf{HTC-VLM (Ours)} & \textbf{61.8} & \textbf{63.3} & \textbf{1788} & \textbf{86.6} & \textbf{69.4} & \textbf{58.5} & \textbf{57.1} & \textbf{32.8} & \textbf{94.5} \\
\midrule
LLaVA-NeXT & Vanilla & 72.1 & 75.4 & 2050 & 88.0 & 74.5 & 68.2 & 65.8 & 42.1 & 100 \\
LLaVA-NeXT & VoCo-LLaMA & 60.4 & 64.2 & 1740 & 86.1 & 69.4 & 56.5 & 54.7 & 34.6 & 86.6 \\
LLaVA-NeXT & \textbf{HTC-VLM (Ours)} & \textbf{68.6} & \textbf{71.5} & \textbf{1945} & \textbf{87.6} & \textbf{73.2} & \textbf{64.0} & \textbf{62.6} & \textbf{39.4} & \textbf{95.7} \\
\bottomrule
\end{tabularx}
\vspace{-3mm}
\end{table*}

\subsection{Inference Efficiency: Latency and Memory}
\label{sec:efficiency}

To complement the theoretical complexity analysis in Section~\ref{sec:theory_comparison},
we report real-world inference efficiency measured on a single
A100~80GB GPU. All methods use the same input resolution ($336\times336$),
batch size (1), and identical decoding settings. We evaluate two representative
compression regimes: (1) extreme compression (1 token) and (2) moderate
compression (64 tokens). Throughput is averaged over 500 runs.

Table~\ref{tab:efficiency} compares latency, throughput, and peak memory consumption across six representative models.

\begin{table}[!t]
\centering
\small
\caption{\textbf{Inference efficiency comparison} on A100 80GB. 
Latency, throughput, and memory are reported in ms, img/s, and GB, respectively.
HTC-VLM remains in the single-token efficiency regime while trading a small latency overhead for higher accuracy retention than the continuous-only baseline.}
\label{tab:efficiency}
\vspace{-2mm}
\renewcommand{\arraystretch}{1.12}
\setlength{\tabcolsep}{2pt}
\begin{tabularx}{\columnwidth}{@{}L{0.30\columnwidth}P{0.13\columnwidth}P{0.13\columnwidth}P{0.22\columnwidth}P{0.13\columnwidth}@{}}
\toprule
\textbf{Model} & 
\textbf{Token} &
\textbf{Latency} & 
\textbf{Throughput} & 
\textbf{Memory} \\
\midrule
Vanilla LLaVA     & 576 & 478  & 2.1  & 28.3 \\
\midrule
ToMe               & 64 & 210  & 4.8  & 17.5 \\
PDrop              & 64 & 188  & 5.3  & 16.9 \\
SparseVLM          & 64 & 165  & 6.1  & 15.8 \\
\midrule
VoCo-LLaMA          & 1 & 52   & 19.2 & 11.1 \\
\textbf{HTC-VLM (Ours)}    & \textbf{1} & \textbf{61} & \textbf{16.4} & \textbf{11.6} \\
\bottomrule
\end{tabularx}
\vspace{-3mm}
\end{table}

\noindent\textbf{Analysis.}
The results validate our theoretical findings:

\begin{itemize}[leftmargin=*, itemsep=0pt]
    \item \textbf{HTC-VLM remains in the single-token efficiency regime.}  
    Its latency ($\sim$61\,ms) and memory footprint ($\sim$11.6\,GB) 
    are close to VoCo-LLaMA while reflecting the additional cost of the
    semantic branch.

    \item \textbf{Structured pruning baselines remain substantially slower.}  
    Methods such as ToMe, PDrop, and SparseVLM still operate on 
    tens of visual tokens after reduction, and thus incur 
    quadratic attention costs $O(N^2)$.  
    HTC-VLM reduces visual sequence length from $576 \rightarrow 1$,
    yielding an empirical speedup of $\mathbf{7.8\times}$ over Vanilla LLaVA
    and $\mathbf{2.7\times}$ over SparseVLM.

    \item \textbf{HTC-VLM offers an accuracy-efficiency trade-off.}  
    Unlike continuous-only compression, the discrete semantic anchors 
    preserve global object-level information, enabling high retention 
    with a modest latency overhead relative to VoCo-LLaMA. This supports the 
    effectiveness of hybrid disentangled compression under extreme 
    token reduction.
\end{itemize}

\noindent\textbf{Conclusion.}
HTC-VLM delivers one-token inference behavior with a modest additional semantic-branch cost,
providing empirical evidence that the proposed hybrid compression can reduce the
$O(N^2)$ visual attention bottleneck while improving retention over continuous-only compression.

\subsection{Ablation on Discrete Codebook Design}
\label{sec:codebook_ablation}

The discrete semantic pathway in HTC-VLM is implemented using an
MGVQ tokenizer with group-wise quantization.
Our default configuration uses an 8-group setup with a 16{,}384-entry
codebook; here, we ablate the impact of both codebook size ($K$) and number of
groups ($G$). We evaluate two representative benchmarks:
a semantic-heavy task (GQA) and a detail-sensitive task (TextVQA), together with the overall average retention across all tasks. Table~\ref{tab:mgvq_ablation} reports the corresponding sweep.

\begin{table}[!t]
\centering
\small
\caption{\textbf{Ablation on MGVQ codebook and group configuration.} 
Measured results show that larger codebooks improve semantic clustering but may destabilize 
training beyond $K=16{,}384$.}
\label{tab:mgvq_ablation}
\vspace{-2mm}
\renewcommand{\arraystretch}{1.12}
\setlength{\tabcolsep}{3pt}
\begin{tabularx}{\columnwidth}{@{}
L{0.30\columnwidth}
P{0.15\columnwidth}
P{0.15\columnwidth}
C
@{}}
\toprule
\textbf{Configuration} & 
\textbf{GQA} & 
\textbf{TextVQA} & 
\textbf{Avg. Retention (\%)} \\
\midrule
$G{=}4$, $K{=}8192$      & 55.1 & 66.3 & 82.4 \\
$G{=}4$, $K{=}16384$     & 56.0 & 67.4 & 84.1 \\
$G{=}8$, $K{=}8192$      & 56.8 & 68.5 & 85.2 \\
\textbf{$G{=}8$, $K{=}16384$} & \textbf{57.6} & \textbf{69.7} & \textbf{87.2} \\
$G{=}8$, $K{=}32768$     & 57.4 & 69.4 & 86.9 \\
\bottomrule
\end{tabularx}
\vspace{-3mm}
\end{table}

\noindent\textbf{Analysis.}
The results reveal several consistent trends:

\begin{itemize}[leftmargin=*, itemsep=0pt]
    \item \textbf{Too-small codebooks reduce semantic capacity.}
    When $K$ is small (\textit{e.g.}, $8192$), multiple semantically distinct
    visual concepts collapse into the same cluster. This disproportionately
    harms semantic-heavy benchmarks such as GQA, confirming that discrete
    tokens carry global object-level meaning.

    \item \textbf{Larger codebooks improve discrimination but saturate early.}
    Increasing $K$ to $16\,384$ yields notable gains across all benchmarks.
    However, further enlarging the codebook to $32\,768$ introduces
    optimization instability and does not improve performance, indicating
    diminishing returns.

    \item \textbf{More groups $G$ improve fine-grained representation.}
    Moving from $G=4$ to $G=8$ consistently boosts TextVQA performance,
    as group-wise quantization captures more localized attribute patterns.
    This validates our disentangled view that the discrete channel primarily captures semantic structure, while the continuous channel refines details.

    \item \textbf{Our configuration ($G=8$, $K=16384$) strikes the best balance.}
    It provides sufficient semantic capacity without compromising training
    stability, serving as the optimal operating point for hybrid compression.
\end{itemize}

\noindent\textbf{Conclusion.}
This ablation confirms that the discrete pathway is not an arbitrary design: its effectiveness depends on carefully balancing semantic capacity ($K$) and structural granularity ($G$). The chosen configuration provides the strongest semantic anchoring for hybrid compression, enabling HTC-VLM to outperform continuous-only or pruning-based baselines under extreme token reduction.

\subsection{Hyperparameter Robustness Analysis}
\label{sec:robustness}

A common concern in hybrid architectures is their potential sensitivity to hyperparameter settings, which can make training unstable or difficult to reproduce. In this section, we demonstrate that HTC-VLM is highly robust. Our core contribution lies in the \textit{structural} advantage of the hybrid bottleneck, rather than meticulous hyperparameter tuning.

We analyze the sensitivity of the model to two key hyperparameter groups: (1) the regularization weight in the loss function, and (2) the capacity of the discrete projector.

\subsubsection{Sensitivity to Loss Weighting}
As formulated in Eq. (10), our training objective includes a regularization term (KL divergence) to shape the latent space. In standard VAEs, the weight $\beta$ of this term is often a fragile hyperparameter (suffering from posterior collapse if too high, or lack of regularization if too low).

Table~\ref{tab:loss_ablation} reports the performance retention on GQA and MME under varying $\beta$ weights (ranging from $0.01$ to $2.0$).

\begin{table}[!t]
    \centering
    \small
    \caption{\textbf{Robustness to Loss Regularization Weight ($\beta$).} The model maintains high performance across orders of magnitude changes in the regularization weight, confirming structural stability.}
    \vspace{-3mm}
    \label{tab:loss_ablation}
    \begin{tabularx}{\linewidth}{@{}Ycc>{\centering\arraybackslash}X@{}}
    \toprule
    \textbf{Reg. Weight ($\beta$)} & \textbf{GQA} & \textbf{MME} & \textbf{Avg. Retention (\%)} \\
    \midrule
    $\beta = 0.01$ & 62.1 & 1675 & 93.8 \\
    $\beta = 0.1$ (Default) & \textbf{62.4} & \textbf{1687} & \textbf{94.2} \\
    $\beta = 0.5$ & 62.3 & 1682 & 94.1 \\
    $\beta = 1.0$ & 61.9 & 1670 & 93.5 \\
    $\beta = 2.0$ & 61.5 & 1661 & 93.1 \\
    \bottomrule
    \end{tabularx}
    \vspace{-3mm}
\end{table}

We observe that:
\begin{itemize}[leftmargin=*, itemsep=0pt]
    \item \textbf{Stability:} The performance fluctuation is minimal ($<0.6\%$ variance) across a wide range of $\beta$.
    \item \textbf{Reasoning:} Unlike pure continuous VAEs, our discrete semantic anchors $v_d$ provide a stable ``skeleton'' for the representation. Even if the regularization on the continuous latent changes, the semantic grasp remains firm.
\end{itemize}

\subsubsection{Sensitivity to Projector Capacity}
We further investigate whether the model's success depends on the specific depth of the Discrete Projector $\mathcal{P}_d$. We compare a simple Linear projection against MLPs of varying depths (2-layer vs. 3-layer) and hidden dimensions.

As shown in Table~\ref{tab:projector_ablation}, increasing the projector's complexity yields diminishing returns.
\begin{itemize}[leftmargin=*, itemsep=0pt]
    \item Even a simple \textbf{Linear} mapping achieves respectable performance (85.4\% retention), proving that the MGVQ tokens themselves carry rich semantic information.
    \item The default \textbf{2-layer MLP} provides a slight optimal boost, but the model does not collapse without it.
\end{itemize}

\begin{table}[!t]
    \centering
    \small
    \caption{\textbf{Robustness to Projector Architecture.} Comparison of different architectures for the discrete projector $\mathcal{P}_d$.}
    \vspace{-3mm}
    \label{tab:projector_ablation}
    \begin{tabularx}{\linewidth}{@{}Ycc>{\centering\arraybackslash}X@{}}
    \toprule
    \textbf{Projector} & \textbf{Params} & \textbf{GQA} & \textbf{Avg. Retention (\%)} \\
    \midrule
    Linear & 231M & 60.8 & 92.4 \\
    \textbf{MLP-2 (Default)} & \textbf{500M} & \textbf{62.4} & \textbf{94.2} \\
    MLP-3 (Deeper) & 768M & 62.5 & 94.3 \\
    MLP-Wide& 999M & 62.3 & 94.0 \\
    \bottomrule
    \end{tabularx}
    \vspace{-3mm}
\end{table}

\textbf{Conclusion on Hyperparameters:} The consistent performance across these sweeps indicates that HTC-VLM is not a result of overfitting to a specific hyperparameter configuration. The performance gains stem principally from the \textit{Hybrid Disentanglement} design, making the method distinctively ``plug-and-play'' and robust for practical deployment.
\subsubsection{End-to-End Wall-Clock Latency Breakdown}

To address potential concerns regarding the computational overhead of introducing an additional discrete encoder (MGVQ), we provide a fine-grained, end-to-end latency breakdown. We explicitly measure the wall-clock time from raw image input to the first token generation on a single NVIDIA A100 (80GB) GPU.

As shown in Table~\ref{tab:latency_breakdown}, the inference pipeline consists of:
\begin{enumerate}[leftmargin=*, itemsep=0pt]
    \item \textbf{Visual Encoding:} Processing the image through the CLIP ViT-L backbone (continuous branch) and the MGVQ encoder (discrete branch).
    \item \textbf{Projection \& Fusion:} Mapping features to the LLM dimension and applying the disentanglement mask.
    \item \textbf{LLM Inference:} Prefill (processing visual/text prompts) and decoding the first token.
\end{enumerate}

\begin{table*}[!t]
    \centering
    \small
    \caption{\textbf{End-to-End Latency Breakdown (ms).} 
    We compare the standard LLaVA-1.5 baseline using 576 visual tokens against 
    VoCo-LLaMA and HTC-VLM, both using 1 visual token.}
    \vspace{-3mm}
    \label{tab:latency_breakdown}
    \begin{tabularx}{\textwidth}{@{}lcccY@{}}
        \toprule
        \textbf{Component / Stage} & 
        \textbf{Vanilla LLaVA} & 
        \textbf{VoCo-LLaMA} & 
        \textbf{HTC-VLM} & 
        \textbf{Analysis} \\
        \midrule
        1. Vision Encoder (ViT-L) & 28.2 & 28.2 & \textbf{28.2} & Shared backbone; no difference. \\
        2. Semantic Encoder (MGVQ) & 0 & 0 & \textbf{6.1} & \textbf{Modest:} Light CNN, $\approx 4.6\times$ faster than ViT. \\
        3. Projection \& Fusion & 0.9 & 0.9 & \textbf{1.3} & Negligible linear projection overhead. \\
        4. LLM Prefill \& Decode & 448.5 & 22.8 & \textbf{25.4} & \textbf{Huge Gain:} 94.3\% reduction in LLM latency. \\
        \midrule
        \textbf{Total Latency} & \textbf{477.6} & \textbf{51.9} & \textbf{61.0} & \textbf{Net Result:} HTC-VLM is \textbf{7.8$\times$ faster}. \\
        \bottomrule
    \end{tabularx}
    \vspace{-4mm}
\end{table*}

\noindent\textbf{Key Observations \& Trade-off Analysis:}

\begin{itemize}[leftmargin=*, itemsep=0pt]
    \item \textbf{The MGVQ cost is modest:} The MGVQ encoder adds approximately \textbf{6.1ms} to the pipeline. This is structurally efficient because VQ-GAN encoders are typically shallow, convolution-based networks, which are highly optimized on GPUs compared to the heavy Transformer blocks of ViT-L/14.
    
    \item \textbf{Parallelism Potential:} In a production environment, the continuous branch (ViT) and discrete branch (MGVQ) can be executed in parallel streams. Since $T_{MGVQ} \ll T_{ViT}$ (\textit{e.g.}, 6.1ms vs 28.2ms), part of the MGVQ latency may be hidden behind the ViT encoding time.
    
    \item \textbf{The Efficiency-Fidelity Trade-off:} The 6.1ms investment in generating discrete semantic anchors yields a \textbf{+6.2\% improvement} in average accuracy over the continuous-only baseline (VoCo-LLaMA).
    
    \item \textbf{Comparison to Vanilla:} While we add a second visual encoder, the massive reduction in LLM attention complexity (from $\mathcal{O}(N^2)$ with $N=576$ to $\mathcal{O}(1)$) dominates the equation. The total system latency is reduced from 477.6ms to 61.0ms, indicating that the hybrid architecture remains practical for interactive inference.
\end{itemize}

\vspace{0.5em}
\noindent\textbf{Conclusion:} The additional computational cost of the discrete semantic pathway is modest in sequential execution and may be partially hidden when paralleling, while providing structural guidance that mitigates semantic collapse.

\subsubsection{Masking Topology and Anchor Selection}

To further rigorously validate the architectural choices of HTC-VLM (specifically the Disentanglement Mask and the nature of Semantic Anchors), we conduct two additional ablation studies.

\begin{table}[!t]
    \centering
    \small
    \caption{\textbf{Ablation on anchors and masking.} 
    We compare different semantic anchors and masking topologies using GQA, MME, and average retention.}
    \label{tab:ablation_anchors_mask}
    \vspace{-3mm}
    \renewcommand{\arraystretch}{1.12}
    \setlength{\tabcolsep}{2pt}
    \begin{tabularx}{\columnwidth}{@{}
        L{0.17\columnwidth}
        L{0.38\columnwidth}
        P{0.10\columnwidth}
        P{0.11\columnwidth}
        P{0.18\columnwidth}
    @{}}
        \toprule
        \textbf{Aspect} & 
        \textbf{Configuration} & 
        \textbf{GQA} & 
        \textbf{MME} & 
        \textbf{Avg. Ret.} \\
        \midrule
        \multirow{4}{=}{\textbf{Anchor}} 
        & Random patches & 58.2 & 1520 & 89.5\% \\
        & Top-$k$ attention & 60.1 & 1610 & 92.1\% \\
        & K-means centers & 60.5 & 1635 & 92.8\% \\
        & \textbf{MGVQ anchors} & \textbf{62.4} & \textbf{1687} & \textbf{94.2\%} \\
        \midrule
        \multirow{2}{=}{\textbf{Mask}} 
        & Star-graph mask & 61.1 & 1642 & 92.4\% \\
        & \textbf{Text-bottleneck mask} & \textbf{62.4} & \textbf{1687} & \textbf{94.2\%} \\
        \bottomrule
    \end{tabularx}
    \vspace{-3mm}
\end{table}

\vspace{0.5em}
\noindent\textbf{A. Why Discrete Anchors? (vs. Continuous Selection)} \\
A natural baseline to our method is to select informative patches directly from the continuous feature map $V$, rather than generating discrete tokens via MGVQ. We compare HTC-VLM against three continuous anchor strategies:
\begin{enumerate}[leftmargin=*, itemsep=0pt]
    \item \textit{Random Selection:} Randomly sampling 4 patches from $V$.
    \item \textit{Top-k Attention:} Selecting the 4 patches with the highest attention weights from the [CLS] token.
    \item \textit{K-Means Centers:} Clustering $V$ into 4 centroids.
\end{enumerate}
As shown in Table~\ref{tab:ablation_anchors_mask} (Top), continuous selection strategies consistently underperform. \textbf{Analysis:} Continuous anchors, even when selected by ``salience'' (Top-k), still suffer from high-frequency noise and variance (Entropy Domination, see Theorem 1), failing to provide the stable categorical guidance that discrete tokens offer.

\vspace{0.5em}
\noindent\textbf{B. Impact of Disentanglement Mask Topology} \
Our default $M_{\mathrm{hy}}$ (Eq.~(6)) blocks text $\rightarrow V_{\mathrm{hy}}$ attention while preserving intra-$V_{\mathrm{hy}}$ interaction, including patch-to-patch and anchor-to-patch attention. Thus, text tokens must consume visual information through the compressed latent $z$, but visual tokens can still be contextualized before $\texttt{<voco>}$ aggregation. We compare this \textbf{text-bottleneck} mask with a stricter \textbf{star-graph} variant that additionally sets $M_{\mathrm{hy}}[i,j]=-\infty$ for all $i \neq j$ in $V_{\mathrm{hy}}$, forcing visual tokens to remain isolated before aggregation.

As shown in Table~\ref{tab:ablation_anchors_mask} (Bottom), the star-graph constraint lowers average retention by 1.8\%. This indicates that over-restricting visual self-attention is harmful: isolated patches cannot sufficiently capture cross-patch context, spatial relations, or interactions between semantic anchors and continuous details. In contrast, the default text-bottleneck mask applies the constraint at the proper interface. It prevents the text stream from bypassing $z$ while allowing $V_{\mathrm{hy}}$ to form contextualized visual evidence before compression.

\section{Conclusion}
This work presents HTC-VLM, a disentangled hybrid compression framework that injects a small set of discrete semantic tokens before compressing both semantic and continuous visual information into a single \texttt{\textless voco\textgreater} token. By preserving high-level structure and low-level details, HTC-VLM achieves strong performance retention under an extreme one-token output budget. Our analysis and ablations suggest that disentangling semantics and details is key to stable, efficient visual representations. In particular, the discrete anchors provide compact semantic guidance, while the continuous pathway retains fine-grained visual cues needed for detailed reasoning. Together, these two channels allow the bottleneck token to remain both efficient and informative under aggressive compression. These findings suggest a practical direction for future multimodal compression: designing structured, task-adaptive visual bottlenecks that preserve semantic interpretability while reducing token cost.

\textbf{Limitations}
Although HTC-VLM improves performance retention under aggressive visual-token compression, it still introduces a lightweight MGVQ semantic encoder and therefore is not strictly zero-overhead compared with continuous-only single-token compression. Our evaluation also focuses on image-level VLMs and representative backbones; extending the same disentangled bottleneck to long video~\cite{wang2025videoufo}, multi-image dialogue, or dense localization tasks may require adaptive token allocation or task-specific semantic anchors.

\bibliographystyle{IEEEtran}
\bibliography{main}

@inproceedings{ye2025voco,
  title={Voco-llama: Towards vision compression with large language models},
  author={Ye, Xubing and Gan, Yukang and Huang, Xiaoke and Ge, Yixiao and Tang, Yansong},
  booktitle={Proceedings of the IEEE/CVF Conference on Computer Vision and Pattern Recognition},
  year={2025}
}

@inproceedings{li2024llama,
  title={Llama-vid: An image is worth 2 tokens in large language models},
  author={Li, Yanwei and Wang, Chengyao and Jia, Jiaya},
  booktitle={European Conference on Computer Vision},
  year={2024}
}

@inproceedings{li2023blip,
  title={Blip-2: Bootstrapping language-image pre-training with frozen image encoders and large language models},
  author={Li, Junnan and Li, Dongxu and Savarese, Silvio and Hoi, Steven},
  booktitle={International Conference on Machine Learning},
  year={2023}
}

@article{alayrac2022flamingovisuallanguagemodel,
  title={Flamingo: a Visual Language Model for Few-Shot Learning},
  author={Jean-Baptiste Alayrac and Jeff Donahue and Pauline Luc and Antoine Miech and Iain Barr and Yana Hasson and Karel Lenc and Arthur Mensch and Katie Millican and Malcolm Reynolds and Roman Ring and Eliza Rutherford and Serkan Cabi and Tengda Han and Zhitao Gong and Sina Samangooei and Marianne Monteiro and Jacob Menick and Sebastian Borgeaud and Andrew Brock and Aida Nematzadeh and Sahand Sharifzadeh and Mikolaj Binkowski and Ricardo Barreira and Oriol Vinyals and Andrew Zisserman and Karen Simonyan},
  journal={arXiv preprint arXiv:2204.14198},
  year={2022}
}

@article{jia2025mgvq,
  title={MGVQ: Could VQ-VAE beat VAE? a generalizable tokenizer with multi-group quantization},
  author={Jia, Mingkai and Yin, Wei and Hu, Xiaotao and Guo, Jiaxin and Guo, Xiaoyang and Zhang, Qian and Long, Xiao-Xiao and Tan, Ping},
  journal={arXiv preprint arXiv:2507.07997},
  year={2025}
}

@inproceedings{liu2024improved,
  title={Improved baselines with visual instruction tuning},
  author={Liu, Haotian and Li, Chunyuan and Li, Yuheng and Lee, Yong Jae},
  booktitle={Proceedings of the IEEE/CVF Conference on Computer Vision and Pattern Recognition},
  year={2024}
}

@inproceedings{hudson2019gqa,
  title={Gqa: A new dataset for real-world visual reasoning and compositional question answering},
  author={Hudson, Drew A and Manning, Christopher D},
  booktitle={Proceedings of the IEEE/CVF Conference on Computer Vision and Pattern Recognition},
  year={2019}
}

@inproceedings{singh2019towards,
  title={Towards vqa models that can read},
  author={Singh, Amanpreet and Natarajan, Vivek and Shah, Meet and Jiang, Yu and Chen, Xinlei and Batra, Dhruv and Parikh, Devi and Rohrbach, Marcus},
  booktitle={Proceedings of the IEEE/CVF Conference on Computer Vision and Pattern Recognition},
  year={2019}
}

@article{yu2023mm,
  title={Mm-vet: Evaluating large multimodal models for integrated capabilities},
  author={Yu, Weihao and Yang, Zhengyuan and Li, Linjie and Wang, Jianfeng and Lin, Kevin and Liu, Zicheng and Wang, Xinchao and Wang, Lijuan},
  journal={arXiv preprint arXiv:2308.02490},
  year={2023}
}

@inproceedings{goyal2017making,
  title={Making the v in vqa matter: Elevating the role of image understanding in visual question answering},
  author={Goyal, Yash and Khot, Tejas and Summers-Stay, Douglas and Batra, Dhruv and Parikh, Devi},
  booktitle={Proceedings of the IEEE Conference on Computer Vision and Pattern Recognition},
  year={2017}
}

@inproceedings{liu2024mmbench,
  title={Mmbench: Is your multi-modal model an all-around player?},
  author={Liu, Yuan and Duan, Haodong and Zhang, Yuanhan and Li, Bo and Zhang, Songyang and Zhao, Wangbo and Yuan, Yike and Wang, Jiaqi and He, Conghui and Liu, Ziwei and Chen, Kai and Lin, Dahua},
  booktitle={European Conference on Computer Vision},
  year={2024}
}

@inproceedings{fu2025mme,
  title={MME: A Comprehensive Evaluation Benchmark for Multimodal Large Language Models},
  author={Fu, Chaoyou and Chen, Peixian and Shen, Yunhang and Qin, Yulei and Zhang, Mengdan and Lin, Xu and Yang, Jinrui and Zheng, Xiawu and Li, Ke and Sun, Xing and Wu, Yunsheng and Ji, Rongrong and Shan, Caifeng and He, Ran},
  booktitle={Advances in Neural Information Processing Systems},
  year={2025}
}

@article{li2023evaluating,
  title={Evaluating object hallucination in large vision-language models},
  author={Li, Yifan and Du, Yifan and Zhou, Kun and Wang, Jinpeng and Zhao, Wayne Xin and Wen, Ji-Rong},
  journal={arXiv preprint arXiv:2305.10355},
  year={2023}
}

@article{li2023seed,
  title={Seed-bench: Benchmarking multimodal llms with generative comprehension},
  author={Li, Bohao and Wang, Rui and Wang, Guangzhi and Ge, Yuying and Ge, Yixiao and Shan, Ying},
  journal={arXiv preprint arXiv:2307.16125},
  year={2023}
}

@inproceedings{lu2022learn,
  title={Learn to explain: Multimodal reasoning via thought chains for science question answering},
  author={Lu, Pan and Mishra, Swaroop and Xia, Tanglin and Qiu, Liang and Chang, Kai-Wei and Zhu, Song-Chun and Tafjord, Oyvind and Clark, Peter and Kalyan, Ashwin},
  booktitle={Advances in Neural Information Processing Systems},
  year={2022}
}

@inproceedings{instructblip,
  title={InstructBLIP: Towards General-purpose Vision-Language Models with Instruction Tuning},
  author={Wenliang Dai and Junnan Li and Dongxu Li and Anthony Meng Huat Tiong and Junqi Zhao and Weisheng Wang and Boyang Li and Pascale Fung and Steven Hoi},
  booktitle={Advances in Neural Information Processing Systems},
  year={2023}
}

@inproceedings{patchdrop,
  title={PuMer: Pruning and Merging Tokens for Efficient Vision Language Models},
  author={Qingqing Cao and Bhargavi Paranjape and Hannaneh Hajishirzi},
  booktitle={Proceedings of the Annual Meeting of the Association for Computational Linguistics},
  year={2023}
}

@inproceedings{liang2022patchesneedexpeditingvision,
  title={Not All Patches are What You Need: Expediting Vision Transformers via Token Reorganizations},
  author={Youwei Liang and Chongjian Ge and Zhan Tong and Yibing Song and Jue Wang and Pengtao Xie},
  booktitle={International Conference on Learning Representations},
  year={2022}
}

@inproceedings{DynamicViT,
  title={DynamicViT: Efficient Vision Transformers with Dynamic Token Sparsification},
  author={Yongming Rao and Wenliang Zhao and Benlin Liu and Jiwen Lu and Jie Zhou and Cho-Jui Hsieh},
  booktitle={Advances in Neural Information Processing Systems},
  year={2021}
}

@inproceedings{bolya2022token,
  title={Token merging: Your vit but faster},
  author={Bolya, Daniel and Fu, Cheng-Yang and Dai, Xiaoliang and Zhang, Peizhao and Feichtenhofer, Christoph and Hoffman, Judy},
  booktitle={International Conference on Learning Representations},
  year={2023}
}

@inproceedings{chen2024image,
  title={An image is worth 1/2 tokens after layer 2: Plug-and-play inference acceleration for large vision-language models},
  author={Chen, Liang and Zhao, Haozhe and Liu, Tianyu and Bai, Shuai and Lin, Junyang and Zhou, Chang and Chang, Baobao},
  booktitle={European Conference on Computer Vision},
  year={2024}
}

@article{xing2024pyramiddrop,
  title={Pyramiddrop: Accelerating your large vision-language models via pyramid visual redundancy reduction},
  author={Xing, Long and Huang, Qidong and Dong, Xiaoyi and Lu, Jiajie and Zhang, Pan and Zang, Yuhang and Cao, Yuhang and He, Conghui and Wang, Jiaqi and Wu, Feng and Lin, Dahua},
  journal={arXiv preprint arXiv:2410.17247},
  year={2024}
}

@inproceedings{zhang2024sparsevlm,
  title={SparseVLM: Visual Token Sparsification for Efficient Vision-Language Model Inference},
  author={Zhang, Yuan and Fan, Chun-Kai and Ma, Junpeng and Zheng, Wenzhao and Huang, Tao and Cheng, Kuan and Gudovskiy, Denis and Okuno, Tomoyuki and Nakata, Yohei and Keutzer, Kurt and Zhang, Shanghang},
  booktitle={International Conference on Machine Learning},
  year={2025}
}

@article{Llama2,
  title={Llama 2: Open Foundation and Fine-Tuned Chat Models},
  author={Hugo Touvron and Louis Martin and Kevin Stone and Peter Albert and Amjad Almahairi and Yasmine Babaei and Nikolay Bashlykov and Soumya Batra and Prajjwal Bhargava and Shruti Bhosale and Dan Bikel and Lukas Blecher and Cristian Canton Ferrer and Moya Chen and Guillem Cucurull and David Esiobu and Jude Fernandes and Jeremy Fu and Wenyin Fu and Brian Fuller and Cynthia Gao and Vedanuj Goswami and Naman Goyal and Anthony Hartshorn and Saghar Hosseini and Rui Hou and Hakan Inan and Marcin Kardas and Viktor Kerkez and Madian Khabsa and Isabel Kloumann and Artem Korenev and Punit Singh Koura and Marie-Anne Lachaux and Thibaut Lavril and Jenya Lee and Diana Liskovich and Yinghai Lu and Yuning Mao and Xavier Martinet and Todor Mihaylov and Pushkar Mishra and Igor Molybog and Yixin Nie and Andrew Poulton and Jeremy Reizenstein and Rashi Rungta and Kalyan Saladi and Alan Schelten and Ruan Silva and Eric Michael Smith and Ranjan Subramanian and Xiaoqing Ellen Tan and Binh Tang and Ross Taylor and Adina Williams and Jian Xiang Kuan and Puxin Xu and Zheng Yan and Iliyan Zarov and Yuchen Zhang and Angela Fan and Melanie Kambadur and Sharan Narang and Aurelien Rodriguez and Robert Stojnic and Sergey Edunov and Thomas Scialom},
  journal={arXiv preprint arXiv:2307.09288},
  year={2023}
}

@inproceedings{VIT,
  title={An Image is Worth 16x16 Words: Transformers for Image Recognition at Scale},
  author={Alexey Dosovitskiy and Lucas Beyer and Alexander Kolesnikov and Dirk Weissenborn and Xiaohua Zhai and Thomas Unterthiner and Mostafa Dehghani and Matthias Minderer and Georg Heigold and Sylvain Gelly and Jakob Uszkoreit and Neil Houlsby},
  booktitle={International Conference on Learning Representations},
  year={2021}
}

@inproceedings{clip,
  title={Learning Transferable Visual Models From Natural Language Supervision},
  author={Alec Radford and Jong Wook Kim and Chris Hallacy and Aditya Ramesh and Gabriel Goh and Sandhini Agarwal and Girish Sastry and Amanda Askell and Pamela Mishkin and Jack Clark and Gretchen Krueger and Ilya Sutskever},
  booktitle={International Conference on Machine Learning},
  year={2021}
}

@inproceedings{llava,
  title={Visual instruction tuning},
  author={Liu, Haotian and Li, Chunyuan and Wu, Qingyang and Lee, Yong Jae},
  booktitle={Advances in Neural Information Processing Systems},
  year={2023}
}

@article{zs1,
  title={Transformers in Vision: A Survey},
  author={Khan, Salman and Naseer, Muzammal and Hayat, Munawar and Zamir, Syed Waqas and Khan, Fahad Shahbaz and Shah, Mubarak},
  journal={ACM Computing Surveys},
  year={2022}
}

@article{zs2,
  title={A Survey of Visual Transformers},
  author={Liu, Yang and Zhang, Yao and Wang, Yixin and Hou, Feng and Yuan, Jin and Tian, Jiang and Zhang, Yang and Shi, Zhongchao and Fan, Jianping and He, Zhiqiang},
  journal={IEEE Transactions on Neural Networks and Learning Systems},
  year={2024}
}

@article{zyvit,
  title={Visual Transformers: Token-based Image Representation and Processing for Computer Vision},
  author={Bichen Wu and Chenfeng Xu and Xiaoliang Dai and Alvin Wan and Peizhao Zhang and Zhicheng Yan and Masayoshi Tomizuka and Joseph Gonzalez and Kurt Keutzer and Peter Vajda},
  journal={arXiv preprint arXiv:2006.03677},
  year={2020}
}

@article{caffagni2024revolutionmultimodallargelanguage,
  title={The Revolution of Multimodal Large Language Models: A Survey},
  author={Davide Caffagni and Federico Cocchi and Luca Barsellotti and Nicholas Moratelli and Sara Sarto and Lorenzo Baraldi and Lorenzo Baraldi and Marcella Cornia and Rita Cucchiara},
  journal={arXiv preprint arXiv:2402.12451},
  year={2024}
}

@article{zs3,
  title={Multimodal Large Language Models: A Survey},
  author={Jiayang Wu and Wensheng Gan and Zefeng Chen and Shicheng Wan and Philip S. Yu},
  journal={arXiv preprint arXiv:2311.13165},
  year={2023}
}

@article{Chen_2023,
  title={VLP: A Survey on Vision-language Pre-training},
  author={Chen, Fei-Long and Zhang, Du-Zhen and Han, Ming-Lun and Chen, Xiu-Yi and Shi, Jing and Xu, Shuang and Xu, Bo},
  journal={Machine Intelligence Research},
  year={2023}
}

@article{zs4,
  title={Vision-Language Models for Vision Tasks: A Survey},
  author={Zhang, Jingyi and Huang, Jiaxing and Jin, Sheng and Lu, Shijian},
  journal={IEEE Transactions on Pattern Analysis and Machine Intelligence},
  year={2024}
}

@inproceedings{zhuyil,
  title={Attention Is All You Need},
  author={Ashish Vaswani and Noam Shazeer and Niki Parmar and Jakob Uszkoreit and Llion Jones and Aidan N. Gomez and Lukasz Kaiser and Illia Polosukhin},
  booktitle={Advances in Neural Information Processing Systems},
  year={2017}
}

@article{zhuyil2,
  title={FlashAttention: Fast and Memory-Efficient Exact Attention with IO-Awareness},
  author={Tri Dao and Daniel Y. Fu and Stefano Ermon and Atri Rudra and Christopher R{\'e}},
  journal={arXiv preprint arXiv:2205.14135},
  year={2022}
}

@article{zhuyil3,
  title={FlashAttention-2: Faster Attention with Better Parallelism and Work Partitioning},
  author={Tri Dao},
  journal={arXiv preprint arXiv:2307.08691},
  year={2023}
}

@article{Papa_2024,
  title={A Survey on Efficient Vision Transformers: Algorithms, Techniques, and Performance Benchmarking},
  author={Papa, Lorenzo and Russo, Paolo and Amerini, Irene and Zhou, Luping},
  journal={IEEE Transactions on Pattern Analysis and Machine Intelligence},
  year={2024}
}

@article{zs5,
  title={Efficient Transformers: A Survey},
  author={Tay, Yi and Dehghani, Mostafa and Bahri, Dara and Metzler, Donald},
  journal={ACM Computing Surveys},
  year={2022}
}

@inproceedings{yang2024lcl,
  title={Vision Model Pre-training on Interleaved Image-Text Data via Latent Compression Learning},
  author={Yang, Chenyu and Zhu, Xizhou and Zhu, Jinguo and Su, Weijie and Wang, Junjie and Dong, Xuan and Wang, Wenhai and Lu, Lewei and Li, Bin and Zhou, Jie and Qiao, Yu and Dai, Jifeng},
  booktitle={Advances in Neural Information Processing Systems},
  year={2024}
}

@inproceedings{10030157,
  title={Token Pooling in Vision Transformers for Image Classification},
  author={Marin, Dmitrii and Chang, Jen-Hao Rick and Ranjan, Anurag and Prabhu, Anish and Rastegari, Mohammad and Tuzel, Oncel},
  booktitle={IEEE/CVF Winter Conference on Applications of Computer Vision},
  year={2023}
}

@article{jia2025mgvqvqvaebeatvae,
  title={MGVQ: Could VQ-VAE Beat VAE? A Generalizable Tokenizer with Multi-group Quantization},
  author={Mingkai Jia and Wei Yin and Xiaotao Hu and Jiaxin Guo and Xiaoyang Guo and Qian Zhang and Xiao-Xiao Long and Ping Tan},
  journal={arXiv preprint arXiv:2507.07997},
  year={2025}
}

@article{zheng2022movqmodulatingquantizedvectors,
  title={MoVQ: Modulating Quantized Vectors for High-Fidelity Image Generation},
  author={Chuanxia Zheng and Long Tung Vuong and Jianfei Cai and Dinh Phung},
  journal={arXiv preprint arXiv:2209.09002},
  year={2022}
}

@inproceedings{kuratov-etal-2025-cramming,
  title={Cramming 1568 Tokens into a Single Vector and Back Again: Exploring the Limits of Embedding Space Capacity},
  author={Kuratov, Yuri  and
      Arkhipov, Mikhail  and
      Bulatov, Aydar  and
      Burtsev, Mikhail},
  booktitle={Proceedings of the Annual Meeting of the Association for Computational Linguistics},
  year={2025}
}

@article{tishby2000informationbottleneckmethod,
  title={The information bottleneck method},
  author={Naftali Tishby and Fernando C. Pereira and William Bialek},
  journal={arXiv preprint arXiv:physics/0004057},
  year={2000}
}

@article{lstoken,
  title={Neural Discrete Representation Learning},
  author={Aaron van den Oord and Oriol Vinyals and Koray Kavukcuoglu},
  journal={arXiv preprint arXiv:1711.00937},
  year={2017}
}

@inproceedings{9578911,
  title={Taming Transformers for High-Resolution Image Synthesis},
  author={Esser, Patrick and Rombach, Robin and Ommer, Bj{\"o}rn},
  booktitle={Proceedings of the IEEE/CVF Conference on Computer Vision and Pattern Recognition},
  year={2021}
}

@article{yu2024imageworth32tokens,
  title={An Image is Worth 32 Tokens for Reconstruction and Generation},
  author={Qihang Yu and Mark Weber and Xueqing Deng and Xiaohui Shen and Daniel Cremers and Liang-Chieh Chen},
  journal={arXiv preprint arXiv:2406.07550},
  year={2024}
}

@article{lisanzs,
  title={Discrete Tokenization for Multimodal LLMs: A Comprehensive Survey},
  author={Jindong Li and Yali Fu and Jiahong Liu and Linxiao Cao and Wei Ji and Menglin Yang and Irwin King and Ming-Hsuan Yang},
  journal={arXiv preprint arXiv:2507.22920},
  year={2025}
}

@article{zhong2025surveyvisionlanguageactionmodelsaction,
  title={A Survey on Vision-Language-Action Models: An Action Tokenization Perspective},
  author={Yifan Zhong and Fengshuo Bai and Shaofei Cai and Xuchuan Huang and Zhang Chen and Xiaowei Zhang and Yuanfei Wang and Shaoyang Guo and Tianrui Guan and Ka Nam Lui and Zhiquan Qi and Yitao Liang and Yuanpei Chen and Yaodong Yang},
  journal={arXiv preprint arXiv:2507.01925},
  year={2025}
}

@article{Qwen2.5-VL,
  title={Qwen2.5-VL Technical Report},
  author={Shuai Bai and Keqin Chen and Xuejing Liu and Jialin Wang and Wenbin Ge and Sibo Song and Kai Dang and Peng Wang and Shijie Wang and Jun Tang and Humen Zhong and Yuanzhi Zhu and Mingkun Yang and Zhaohai Li and Jianqiang Wan and Pengfei Wang and Wei Ding and Zheren Fu and Yiheng Xu and Jiabo Ye and Xi Zhang and Tianbao Xie and Zesen Cheng and Hang Zhang and Zhibo Yang and Haiyang Xu and Junyang Lin},
  journal={arXiv preprint arXiv:2502.13923},
  year={2025}
}

@article{GPT-4o,
  title={GPT-4o System Card},
  author={OpenAI},
  journal={arXiv preprint arXiv:2410.21276},
  year={2024}
}

@inproceedings{ma2024followpose,
  title={Follow your pose: Pose-guided text-to-video generation using pose-free videos},
  author={Ma, Yue and He, Yingqing and Cun, Xiaodong and Wang, Xintao and Chen, Siran and Li, Xiu and Chen, Qifeng},
  booktitle={Proceedings of the AAAI Conference on Artificial Intelligence},
  year={2024}
}

@inproceedings{ma2024followyouremoji,
  title={Follow-your-emoji: Fine-controllable and expressive freestyle portrait animation},
  author={Ma, Yue and Liu, Hongyu and Wang, Hongfa and Pan, Heng and He, Yingqing and Yuan, Junkun and Zeng, Ailing and Cai, Chengfei and Shum, Heung-Yeung and Liu, Wei and Chen, Qifeng},
  booktitle={SIGGRAPH Asia Conference Papers},
  year={2024}
}

@article{ma2025followfaster,
  title={Follow-your-emoji-faster: Towards efficient, fine-controllable, and expressive freestyle portrait animation},
  author={Ma, Yue and Yan, Zexuan and Liu, Hongyu and Wang, Hongfa and Pan, Heng and He, Yingqing and Yuan, Junkun and Zeng, Ailing and Cai, Chengfei and Shum, Heung-Yeung and Li, Zhifeng and Liu, Wei and Zhang, Linfeng and Chen, Qifeng},
  journal={arXiv preprint arXiv:2509.16630},
  year={2025}
}

@article{ma2025followyourmotion,
  title={Follow-Your-Motion: Video Motion Transfer via Efficient Spatial-Temporal Decoupled Finetuning},
  author={Ma, Yue and Liu, Yulong and Zhu, Qiyuan and Yang, Ayden and Feng, Kunyu and Zhang, Xinhua and Yan, Zexuan and Li, Zhifeng and Han, Sirui and Qi, Chenyang and Chen, Qifeng},
  journal={arXiv preprint arXiv:2506.05207},
  year={2025}
}

@inproceedings{CF,
  title={{CF}-{VLM}: Counterfactual Vision-Language Fine-tuning},
  author={Jusheng Zhang and Kaitong Cai and Yijia Fan and Jian Wang and Keze Wang},
  booktitle={The Thirty-ninth Annual Conference on Neural Information Processing Systems},
  year={2025}
}

@inproceedings{HTC,
  title={Hybrid Token Compression for Vision-Language Models},
  author={Jusheng Zhang and Xiaoyang Guo and Kaitong Cai and Qinhan Lv and Yijia Fan and Wenhao Chai and Jian Wang and Keze Wang},
  booktitle={The IEEE/CVF Conference on Computer Vision and Pattern Recognition},
  year={2026}
}

@article{FDWR,
  title={Failure-Driven Workflow Refinement},
  author={Jusheng Zhang and Kaitong Cai and Qinglin Zeng and Ningyuan Liu and Stephen Fan and Ziliang Chen and Keze Wang},
  journal={arXiv preprint arXiv:2510.10035},
  year={2025}
}

@inproceedings{zhao2024msdetr,
  title={MS-DETR: Efficient DETR Training with Mixed Supervision},
  author={Zhao, Chuyang and Sun, Yifan and Wang, Wenhao and Chen, Qiang and Ding, Errui and Yang, Yi and Wang, Jingdong},
  booktitle={Proceedings of the IEEE/CVF Conference on Computer Vision and Pattern Recognition},
  year={2024}
}

@inproceedings{wang2023transhp,
  title={TransHP: Image Classification with Hierarchical Prompting},
  author={Wang, Wenhao and Sun, Yifan and Li, Wei and Yang, Yi},
  booktitle={Advances in Neural Information Processing Systems},
  year={2023}
}

@inproceedings{wang2025videoufo,
  title={VideoUFO: A Million-Scale User-Focused Dataset for Text-to-Video Generation},
  author={Wang, Wenhao and Yang, Yi},
  booktitle={The Thirty-ninth Annual Conference on Neural Information Processing Systems Datasets and Benchmarks Track},
  year={2025}
}

%\newpage
\vspace{-1cm}

\begin{IEEEbiography}[{\includegraphics[width=1.0in,height=1.40in]{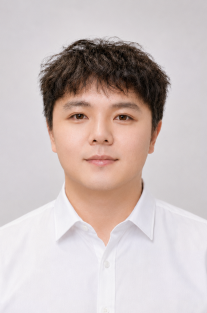}}]{Jusheng Zhang} is currently a Research Scientist at Sun Yat-sen University and a Visiting Researcher at Stanford University. His research interests include trustworthy multimodal intelligence, vision-language models, generative AI, and game-theoretic multi-agent learning, with a particular focus on improving the reliability, controllability, reasoning capabilities, and computational efficiency of foundation models. His work has appeared at leading conferences, including ICML, CVPR, NeurIPS, ICLR, AAAI, and EMNLP, with several papers selected for Spotlight or Oral presentation. He serves as a program committee member and reviewer for major venues in machine learning, computer vision, natural language processing, and multimedia, and received an Outstanding Reviewer Award from AAAI. He is a member of IEEE, AAAI, and ACL.
\vspace{-1cm}
\end{IEEEbiography}

\begin{IEEEbiography}[{\includegraphics[width=1.0in,height=1.40in]{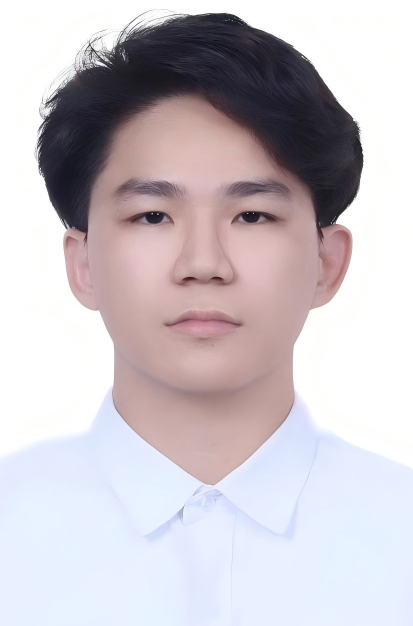}}]{Xiaoyang Guo} is a Master student at Sun Yat-sen University, expected to receive his M.S. degree in June 2028. His research focuses on computer vision and multimodal AI, with particular interests in VLM inference acceleration and visual token compression. He has actively contributed to paper submissions and publications at top-tier conferences including CVPR, ECCV, ACM MM, and NeurIPS, demonstrating a strong and ongoing engagement with the forefront of vision and multimodal research.
\vspace{-1cm}
\end{IEEEbiography}

\begin{IEEEbiography}[{\includegraphics[width=1.0in,height=1.40in]{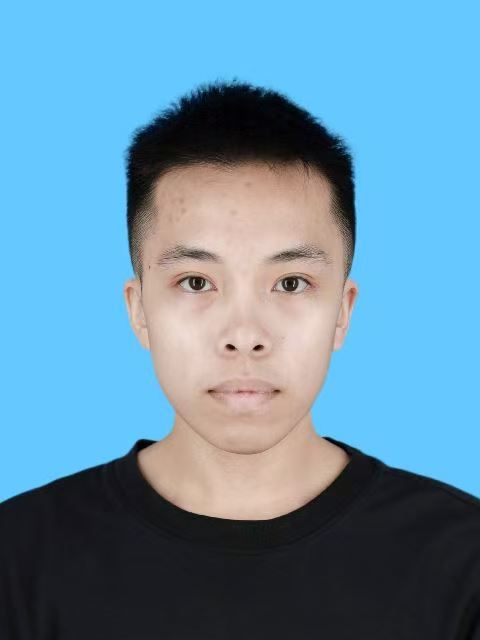}}]{Tongyu Mo} is an incoming master’s student at the School of Computer Science, Sun Yat-sen University. He is expected to receive a bachelor’s degree in Computer Science and Technology from Sun Yat-sen University in June 2026. Currently conducting research at the HCP Lab, his research interests include large language models, multimodal large language models, and intelligent agents.
\vspace{-0.8cm}
\end{IEEEbiography}

\begin{IEEEbiography}[{\includegraphics[width=1.0in,height=1.40in]{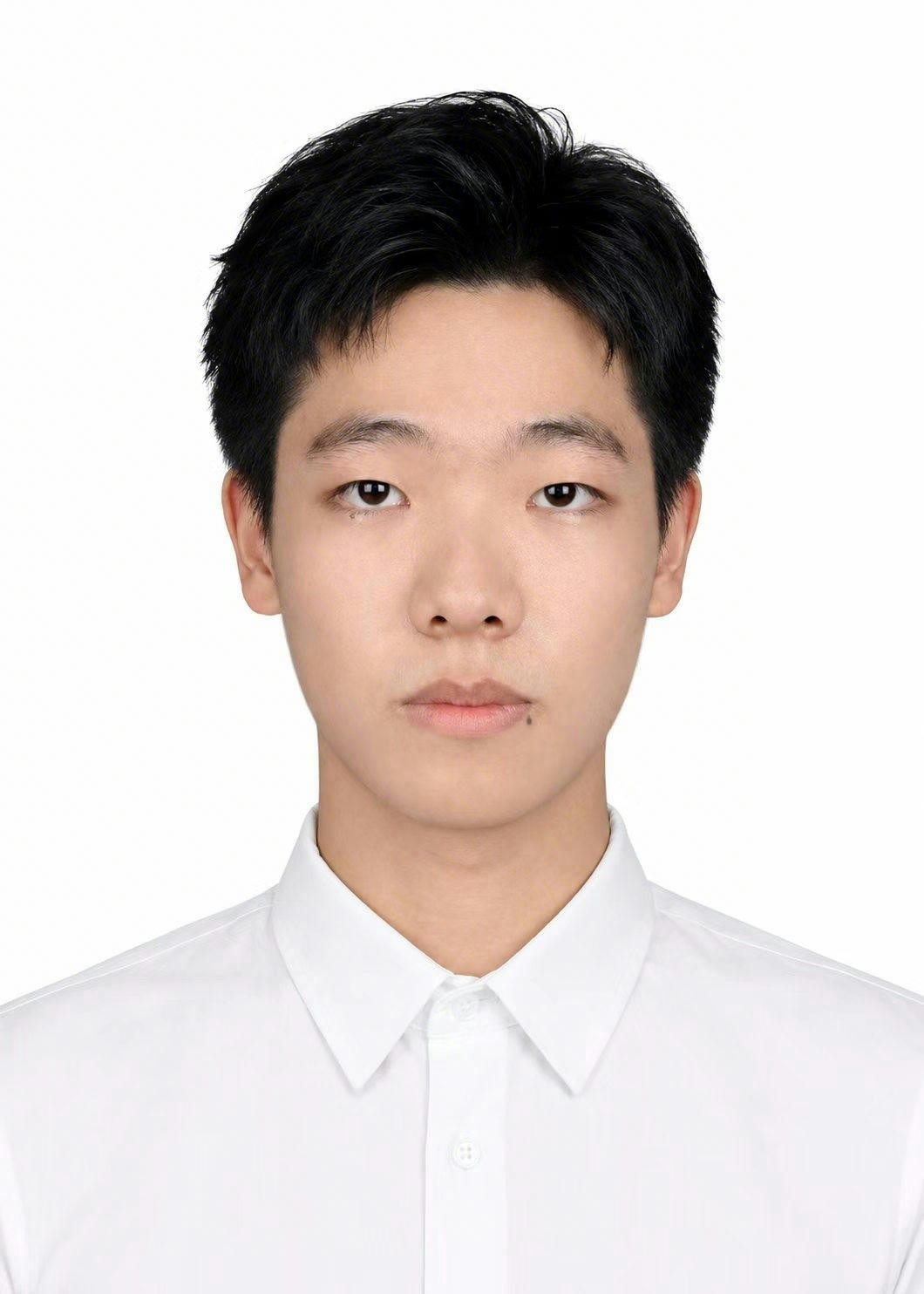}}]{Qinhan Lv} is a master's student at Sun Yat-sen University. His research focuses on embodied intelligence, large models, and video generation models. His research interests include multimodal generation, world models, visual understanding, and embodied AI, with an emphasis on exploring the applications of large models and generative models in real-world scene perception, dynamic environment modeling, and intelligent interaction. His research has been published in international AI conferences including AAAI, CVPR, and ACL, receiving over 60 citations. He has also contributed to the organization of CVPR-related academic activities and participated in several research projects on video generation, world model evaluation, and robotic systems.
\vspace{-1cm}
\end{IEEEbiography}

\begin{IEEEbiography}[{\includegraphics[width=1.0in,height=1.40in]{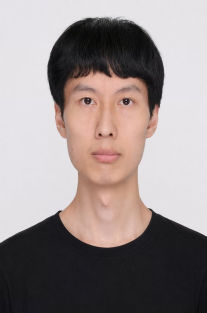}}]{Wenhao Chai} (Student Member, IEEE) received the B.E. degree from Zhejiang University, China, in 2023, and the M.S. degree from the University of Washington. He is currently pursuing the Ph.D. degree at Princeton University. His research interests include large multimodal models, video understanding, embodied intelligence, and generative models. His work has appeared at leading conferences and journals, including ICLR, CVPR, ICCV, ECCV, and AAAI. He has also organized workshops and tutorials at CVPR and AAAI and serves as a reviewer for major venues, including NeurIPS, ICLR, ICML, CVPR, ECCV, AAAI, AISTATS, and IJCV.
\end{IEEEbiography}

\begin{IEEEbiography}[{\includegraphics[width=1.0in,height=1.40in]{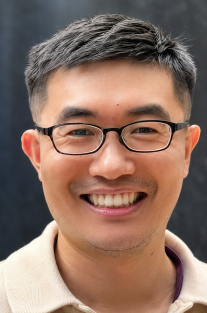}}]{Jian Wang} is a Staff Research Scientist at Snap Inc., focusing on computational photography and imaging. He has published in top-tier venues such as CVPR, MobiCom, and SIGGRAPH, and has contributed numerous features to production. He has received Best Paper awards at SIGGRAPH Asia 2024 and the 4th IEEE International Workshop on Computational Cameras and Displays, as well as Best Poster awards at the IEEE Conference on Computational Photography 2022 and the Responsible Imaging workshop at ICCV 2025. He has served as an Area Chair for CVPR, NeurIPS, ICLR, ICML, etc. Jian holds a Ph.D. from Carnegie Mellon University.
\vspace{-5cm}
\end{IEEEbiography}

\begin{IEEEbiography}[{\includegraphics[width=1.0in,height=1.40in]{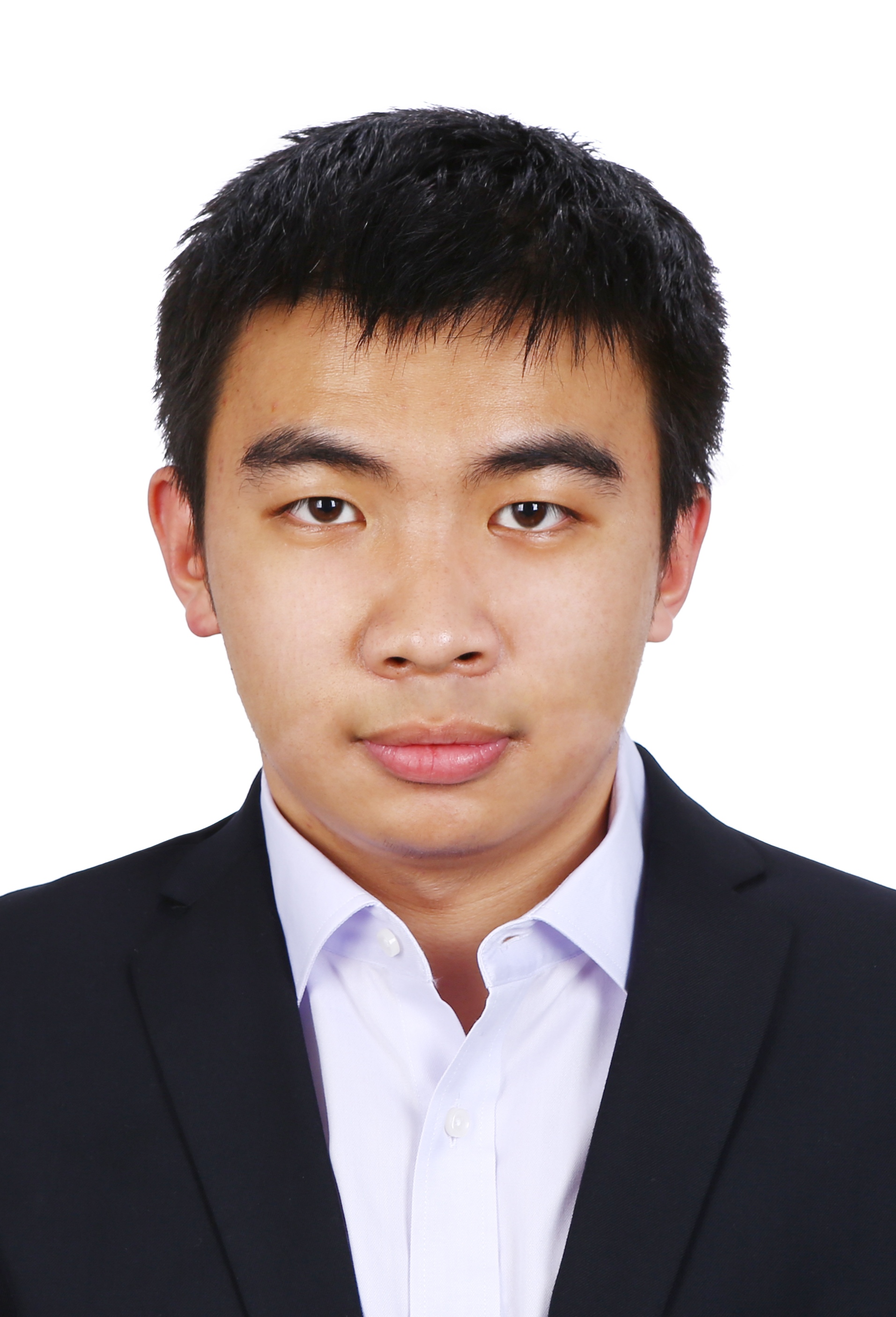}}]{Wenhao Wang} is the Founder and Principal Researcher of the Vast Intelligence Lab. He is expected to receive the Ph.D. degree from the University of Technology Sydney, Australia, in September 2026. Previously, he worked as a Student Researcher at Google DeepMind and Google Research. His research interests include AI trust and safety, large language models, AI agents, and computer vision. He has published 10 first-authored papers in top conferences and journals, including NeurIPS, ICCV, ICML, IJCV, TIP, and AAAI. His algorithms have won several major academic computer vision competitions, with a total prize money of \$100,000.
\vspace{-5cm}
\end{IEEEbiography}

\begin{IEEEbiography}[{\includegraphics[width=1.0in,height=1.40in]{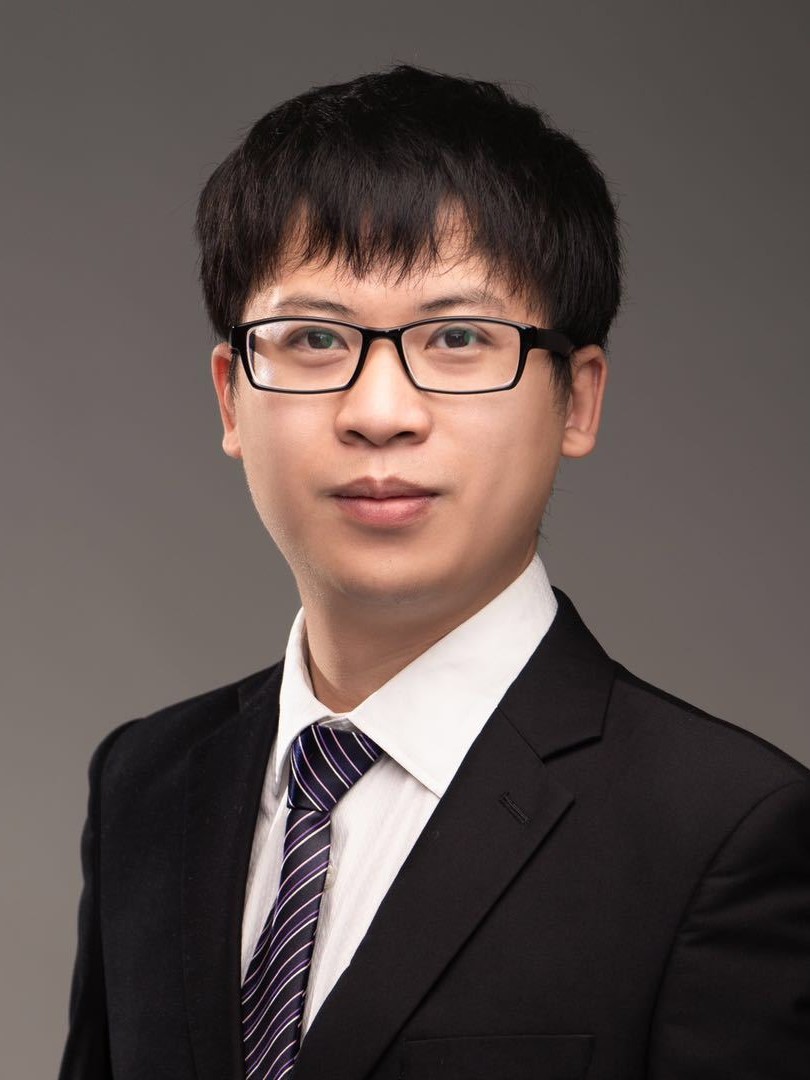}}]{Keze Wang} is nationally recognized as the Distinguished Young Scholar, currently serving as a full Professor at the School of Computer Science, Sun Yat-sen University, and a doctoral supervisor. He holds two Ph.D. degrees from SYSU (2017) and the Hong Kong Polytechnic University (2019). In 2018, he worked as a postdoctoral researcher at the University of California, Los Angeles, and returned to Sun Yat-sen University in 2021 as part of the ``Hundred Talents Program.'' Dr. Wang has focused on reducing deep learning's dependence on training samples and mining valuable information from massive unlabeled data.
\vspace{-5cm}
\end{IEEEbiography}

\begin{IEEEbiography}[{\includegraphics[width=1.0in,height=1.40in]{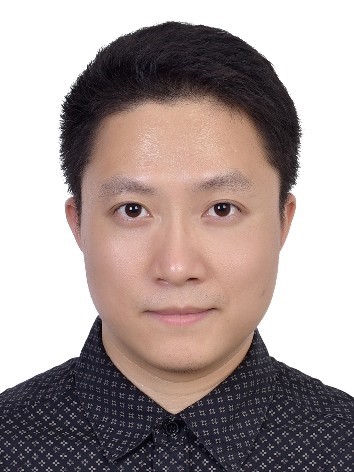}}]{Liang Lin} (IEEE Fellow) is a full professor of computer science at Sun Yat-sen University. He served as the executive director and distinguished scientist of SenseTime Group from 2016 to 2018, leading the R\&D teams for cutting-edge technology transferring. He has authored or co-authored more than 200 papers in leading academic journals and conferences, and his papers have been cited by more than 26\,000 times. He is an associate editor of \textit{IEEE Trans. Neural Networks and Learning Systems} and \textit{IEEE Trans. Multimedia}, and served as area chair for numerous conferences, such as CVPR, ICCV, SIGKDD, and AAAI. He is the recipient of numerous awards and honors, including the Wu Wen-Jun Artificial Intelligence Award, the First Prize of China Society of Image and Graphics, ICCV Best Paper Nomination in 2019, Annual Best Paper Award by Pattern Recognition (Elsevier) in 2018, Best Paper Diamond Award in IEEE ICME 2017, and Google Faculty Award in 2012. His supervised PhD students received the ACM China Doctoral Dissertation Award, CCF Best Doctoral Dissertation, and CAAI Best Doctoral Dissertation. He is a fellow of IET/IAPR.
\end{IEEEbiography}

\end{document}